\documentclass[sigconf]{acmart}
\usepackage{amssymb}
\usepackage{amsfonts}
\usepackage{graphicx}
\usepackage{xcolor}
\usepackage{makecell}
\usepackage{booktabs} 
\usepackage{subfig}
\usepackage{amsmath}
\usepackage{accents}
\usepackage{statex}
\usepackage[normalem]{ulem}
\usepackage{enumitem}
\usepackage{multirow}
\usepackage[nomargin,inline,marginclue,draft]{fixme}
\usepackage{balance}
\usepackage{changepage}
\usepackage{bm}
\usepackage{microtype}
\usepackage{setspace}
\usepackage{hyperref}
\usepackage{mathrsfs}
\usepackage{ulem}
\usepackage{verbatim}
\usepackage{diagbox}
\usepackage{pdftexcmds}
\usepackage{catchfile}
\usepackage{ifluatex}
\usepackage{ifplatform}
\usepackage{threeparttable}

\usepackage{comment}

\usepackage{algorithm}
\usepackage{algorithmicx}
\usepackage{algpseudocode}
\usepackage{amsmath}

\newlength\savedwidth

\author{Jinzhu~Mao}
\affiliation{
\institution{Department of Electronic Engineering, BNRist, \\Tsinghua University, \\Beijing, China}
}
\email{maojz22@mails.tsinghua.edu.cn}

\author{Liu~Cao}
\affiliation{
\institution{Department of Electronic Engineering, BNRist, \\Tsinghua University, \\Beijing, China}
}
\email{l-cao20@mails.tsinghua.edu.cn}

\author{Chen~Gao$^{*}$}
\affiliation{
\institution{Department of Electronic Engineering, BNRist, \\Tsinghua University, \\Beijing, China}
}
\email{chgao96@gmail.com}

\author{Huandong~Wang}
\affiliation{
\institution{Department of Electronic Engineering, BNRist, \\Tsinghua University, \\Beijing, China}
}
\email{wanghuandong@tsinghua.edu.cn}

\author{Hangyu~Fan}
\affiliation{
\institution{Tsingroc Inc.,\\Beijing, China}
}
\email{fanhangyu@tsingroc.com}

\author{Depeng~Jin}
\affiliation{
\institution{Department of Electronic Engineering, BNRist, \\Tsinghua University, \\Beijing, China}
}
\email{jindp@tsinghua.edu.cn}

\author{Yong~Li}
\affiliation{
\institution{Department of Electronic Engineering, BNRist, \\Tsinghua University, \\Beijing, China}
}
\email{liyong07@tsinghua.edu.cn}
\thanks{The first two authors contribute equally to this work. \\ Chen Gao$^{*}$ is the corresponding author. }

\acmISBN{979-8-4007-0103-0/23/08}

\renewcommand{\shortauthors}{Jinzhu Mao et al.}
\copyrightyear{2023}
\acmYear{2023}
\setcopyright{rightsretained}
\acmConference[KDD '23]{Proceedings of the 29th ACM SIGKDD Conference on Knowledge Discovery and Data Mining}{August 6--10, 2023}{Long Beach, CA, USA}
\acmBooktitle{Proceedings of the 29th ACM SIGKDD Conference on Knowledge Discovery and Data Mining (KDD '23), August 6--10, 2023, Long Beach, CA, USA}
\acmDOI{10.1145/3580305.3599804}
\acmISBN{979-8-4007-0103-0/23/08}

\renewcommand\footnotetextcopyrightpermission[1]{} 

\makeatletter
\gdef\@copyrightpermission{
  \begin{minipage}{0.3\columnwidth}
   \href{https://creativecommons.org/licenses/by/4.0/}{\includegraphics[width=0.90\textwidth]{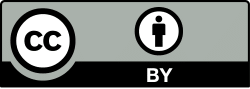}}
  \end{minipage}\hfill
  \begin{minipage}{0.7\columnwidth}
   \href{https://creativecommons.org/licenses/by/4.0/}{This work is licensed under a Creative Commons Attribution International 4.0 License.}
  \end{minipage}
  \vspace{5pt}
}
\makeatother
\begin{document}
\title{Detecting Vulnerable Nodes in Urban Infrastructure Interdependent Network}

\begin{abstract}
	Understanding and characterizing the vulnerability of urban infrastructures, which refers to the engineering facilities essential for the regular running of cities and that exist naturally in the form of networks, is of great value to us. Potential applications include protecting fragile facilities and designing robust topologies, etc. Due to the strong correlation between different topological characteristics and infrastructure vulnerability and their complicated evolution mechanisms, some heuristic and machine-assisted analysis fall short in addressing such a scenario. In this paper, we model the interdependent network as a heterogeneous graph and propose a system based on graph neural network with reinforcement learning, which can be trained on real-world data, to characterize the vulnerability of the city system accurately. The presented system leverages deep learning techniques to understand and analyze the heterogeneous graph, which enables us to capture the risk of cascade failure and discover vulnerable infrastructures of cities. Extensive experiments with various requests demonstrate not only the expressive power of our system but also transferring ability and necessity of the specific components. All source codes and models including those that can reproduce all figures analyzed in this work are publicly available at this link: \url{https://github.com/tsinghua-fib-lab/KDD2023-ID546-UrbanInfra}.
\end{abstract}

\ccsdesc[500]{Information systems~Information systems applications}

\ccsdesc[500]{Information systems~Data mining}

\keywords{Urban Infrastructure Network; Interdependent Network; Graph Neural Networks; Reinforcement Learning}
\maketitle
	
\section{Introduction}\label{sec::intro}

\begin{figure*}[t]
  \centering
  \includegraphics[width=17cm]{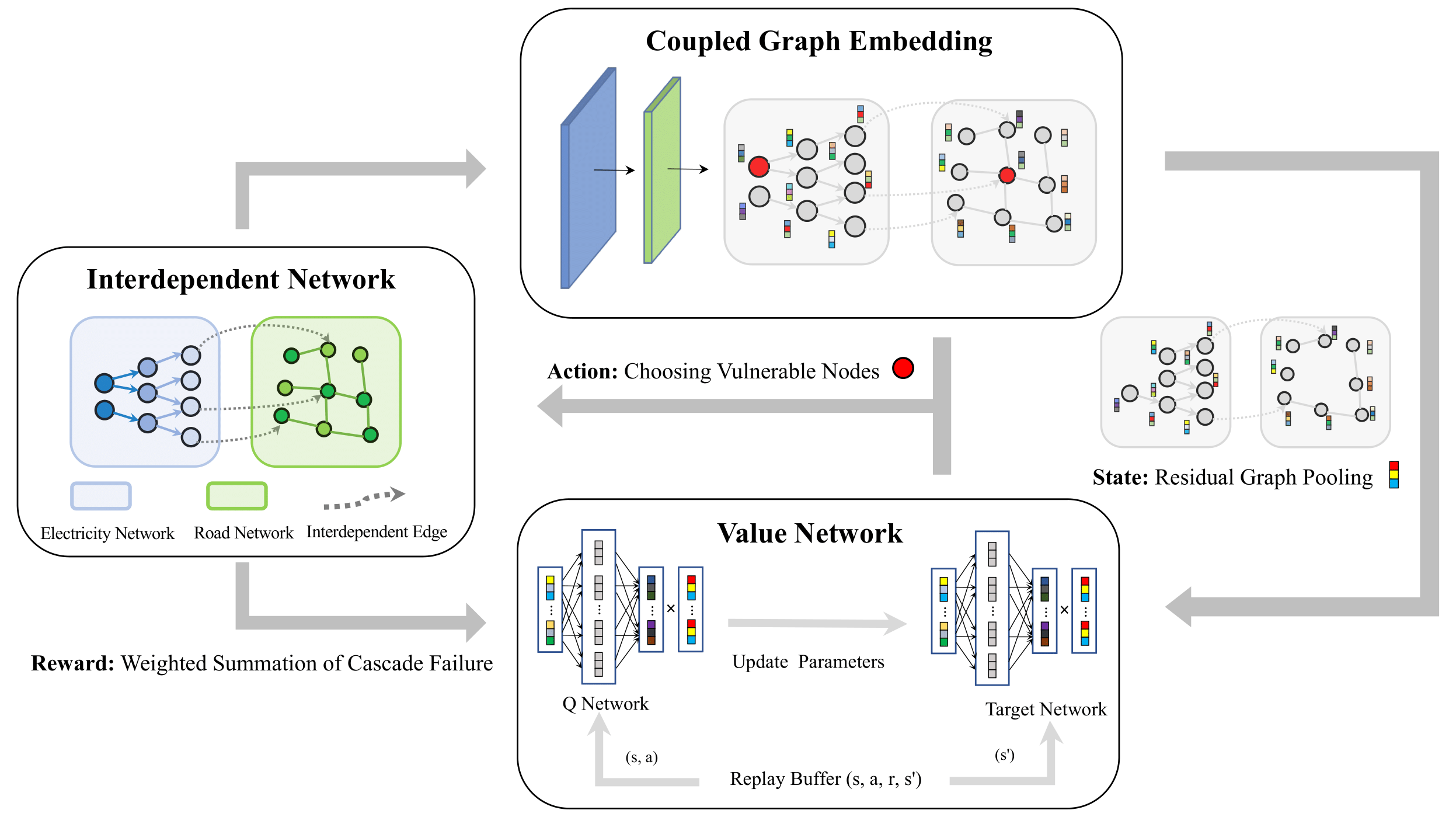}
  \setlength{\abovecaptionskip}{0cm}
        \setlength{\belowcaptionskip}{0.2cm}
  \caption{Illustration of our proposed system.} 
  \label{fig::framework}
\end{figure*}

Urban infrastructure \cite{collier2016urban} refers to the engineering facilities essential for the regular running of cities, including electricity supply, transportation, and communication, etc.
These urban infrastructures exist naturally in the form of networks, including electricity networks, road networks, communication networks, etc.
Further, various infrastructure networks are coupled and interdependent, forming a typical system of interdependent networks \cite{buldyrev2010catastrophic,brummitt2012suppressing}.
Moreover, cities confront the danger of natural catastrophes such as earthquakes, typhoons, and rainstorms, which often destroy functional units of different urban infrastructures.
Due to the interplay between different infrastructures and their functional reliance, destruction often extends beyond the affected area, leading to widespread paralysis of urban infrastructure. This exacerbates the vulnerability of urban infrastructures \cite{buldyrev2010catastrophic,brummitt2012suppressing}.
Urban infrastructures are essential for the proper functioning of various services in a city, including housing and production, and are closely tied to the well-being of its citizens. Therefore, gaining a profound understanding of the vulnerability of urban infrastructure holds significant value.

However, understanding and characterizing the vulnerability of urban infrastructure is a difficult task.
First, networks of different urban infrastructures exhibit diverse topological structures \cite{dunn2016spatial,eusgeld2009role,ren2022topological}, which differ in terms of cluster structure, cycle structure, degree distribution, centrality, etc. The high-dimensional topological structure and the diverse topological features make it challenging for knowledge-based expert systems~\cite{hines2010topological} or feature engineering methods~\cite{abdulla2020predicting} to thoroughly characterize it, which is the first challenge.
Second, there is a strong correlation between these topological characteristics and infrastructure vulnerability \cite{gao2016universal,cohen2000resilience}.
The states of different infrastructure networks have complicated evolution mechanisms. For example, the evolution of vehicle distribution on road networks is driven by the interaction between traffic control signals and driver actions, while the evolution of the distribution of power flow in electricity networks is driven by Ohm's law. These evolution mechanisms compound the functional dependencies of the infrastructure network to exacerbate the higher-order propagation effects of damage. How to model the complicated effect of evolution mechanisms of urban infrastructures is the second challenge.

In this paper, we focus on modeling the vulnerability of interdependent networks composed of diverse urban infrastructures using a data-driven approach.
We develop a graph convolutional neural network, which can efficiently and exhaustively extract the topological features of each infrastructure network as well as the topological features of the interdependency between them, via optimizing the loss function of graph reconstruction.
Further, we construct a simulator of the interdependent networks composed of urban infrastructures and develop a reinforcement learning 
(RL) module by interacting with it. This module deeply models the network evolution mechanism by using the influence of damaged functional units as a reward, which can accurately characterize the vulnerability of the system.

The contributions of this work can be summarized as follows:
\begin{itemize}[leftmargin=*]
    \item We propose a graph convolutional neural network, which performs a coupled graph construction through semi-supervised embedding learning and can efficiently and exhaustively extract the topological features of each infrastructure network as well as the interdependency between them.
    \item We construct a reinforcement learning (RL) module, which makes full use of diverse graph embeddings. It can discover a set of vulnerable functional units of infrastructures in the specified reward setting, thus accurately characterizing the vulnerability of the system. 
    \item Extensive experimental results show that our proposed model can effectively characterize the vulnerability of the system while interacting with an urban simulator. Specifically, the influence of the damaged functional units selected by our proposed algorithm significantly beat those of state-of-the-art algorithms.
\end{itemize}

\section{Problem Statement}\label{sec::profdef}

Starting from the basic interdependent network, we select two closely connected infrastructure networks -- road network \cite{darong2015vulnerability} and electricity network \cite{ren2016cascade} from the city. 
The road network consists of a multitude of roads and intersections, representing edges and nodes, respectively. and the  electricity network is a tree-like structure digraph composed of a large number of power stations at different levels, such as 220 kV, 110 kV, and 10 kV power stations. These two networks are connected by traffic lights located at intersections and low-level (10kV) power stations. 
In the road network, the state of traffic lights directly affects the connectivity of the road network. Specifically, if a traffic light stops working due to damage or power failure, roads it connects will become chaotic and even crowded, and more roads could be affected over time. 
As for the electricity network, it transmits electricity level by level to infrastructures of other networks. 
For instance, the failure of a 220 kV power station will cascade to 110 kV power stations, resulting in the shutdown of certain 10 kV power stations, ultimately impacting the functioning of traffic lights linked to those 10 kV power stations.
Therefore, there are three states for each infrastructure: normal, damaged, and invalid, and "invalid" means that a node stops working because the nodes it connected are damaged or invalid.

To represent the interdependent network, we consider a coupled graph $\mathcal{G}=(\mathcal{V}, \mathcal{E})$ that consists two different structural graphs: the road network $\mathcal{G}^{r}=(\mathcal{V}^{r}, \mathcal{E}^{r})$ and the electricity network $\mathcal{G}^{e}=(\mathcal{V}^{e}, \mathcal{E}^{e})$. 
$\mathcal{V} = \mathcal{V}^{r} \cup \mathcal{V}^{e}$ is the set of nodes, $\mathcal{E} = \mathcal{E}^{r} \cup \mathcal{E}^{e} \cup \mathcal{E}^{'}$ is the set of edges, where $\mathcal{E}^{'}=\left\{e_{v^{e} v^{r}}, \forall v^e \in \mathcal{V}^e, \forall v^r \in \mathcal{V}^r\right\}$ is the set of interdependent directed edges between power stations and traffic lights. A notation table is organized to make this paper easier to understand shown in Table \ref{tabel:notation}. We expect to discover a set of nodes from the interdependent network as the vulnerable nodes, that once they are damaged, the coupled graph will be greatly affected, where "affected" depends on scene and purpose when making decisions. 
For instance, if we pay more attention to the condition of road network, the number of crowded roads can serve as the primary indicator of "affected"; if we care more about the state of electricity network, the power decrease in the grid could be the main indicator of "affected". 
\begin{table}[t]

\centering

\vspace{-0.1cm}
\caption{Notations.} 
\label{tab:ablation}
\vspace{-0.3cm}
\resizebox{0.9\columnwidth}{!}{
\begin{tabular}{c|l}
\toprule
Symbols & Description\\
\midrule
$\mathcal{G}=(\mathcal{V}, \mathcal{E})$ & Coupled graph $\mathcal{G}$, where $\mathcal{V}$ is the set of nodes and $\mathcal{E}$ is the set of edges\\
\midrule
$\mathcal{G}^{e}=(\mathcal{V}^{e}, \mathcal{E}^{e})$ & Electric network $\mathcal{G}^{e}$, where $\mathcal{V}^{e}$ is the set of power stations and $\mathcal{E}^{e}$ is the set of wires\\
\midrule
$\mathcal{E}^{'}$ & Set of interdependent directed edges\\
\midrule
$\mathbf{f}_v$, $\mathbf{h}_v^i$, $\mathbf{f}_v$ & Initial, process and final embedding of node $v$ \\
\midrule
$\mathcal{N}(\cdot)$ & Set of neighbor nodes of given one \\
\midrule
$L$ & Iteration number or depth of GNN \\
\midrule
$\mathbf{W}^{i}$ & Weight matrix of graph neural network at depth $i$ \\
\midrule
$\mathcal{G}^{n}$, $\mathcal{G}^{p}$ & Negative graph and positive graph \\
\midrule
$\mathcal{S}$ & Set of states \\
\midrule
$\mathcal{A}$ & Set of actions \\
\midrule
$\mathcal{P}(s^{\prime}\mid s, a)$ & Transition function, where $s^{\prime}$ is the state at next step \\
\midrule
$\mathcal{R}\left(s, a\right)$ & Reward function \\
\midrule
$\gamma$ & Discount factor \\
\midrule
$v_k$,$s_k$, $r_k$ & Selected node, state and reward at step $k$ \\
\midrule
$Q(s, v)$ & Action-value function \\
\midrule
$P(v^e, \mathcal{G}^e)$ & Function that can calculate the decreased power of $\mathcal{G}^e$ after $v^e$ is damaged \\
\midrule
$E(v^e, \mathcal{G})$ & Function that returns a set of invalid traffic lights affected by damaged $v^{e}$ \\
\midrule
$A(v^r, \mathcal{G}^r)$ & Function that can calculate the decreased value of connectivity of $\mathcal{G}^r$ after $v^r$ is damaged \\
\midrule
$a^e$, $a^r$ & Weight coefficients of $\mathcal{G}^e$ and $\mathcal{G}^r$ \\
\bottomrule
\end{tabular}}
\label{tabel:notation}

\vspace{-0.3cm}
\end{table}

\section{The proposed system}\label{sec::method}

\begin{figure*}[t]
  \centering
  \subfloat[Power Decrease of Electricity Network]{\includegraphics[width=5.5cm]{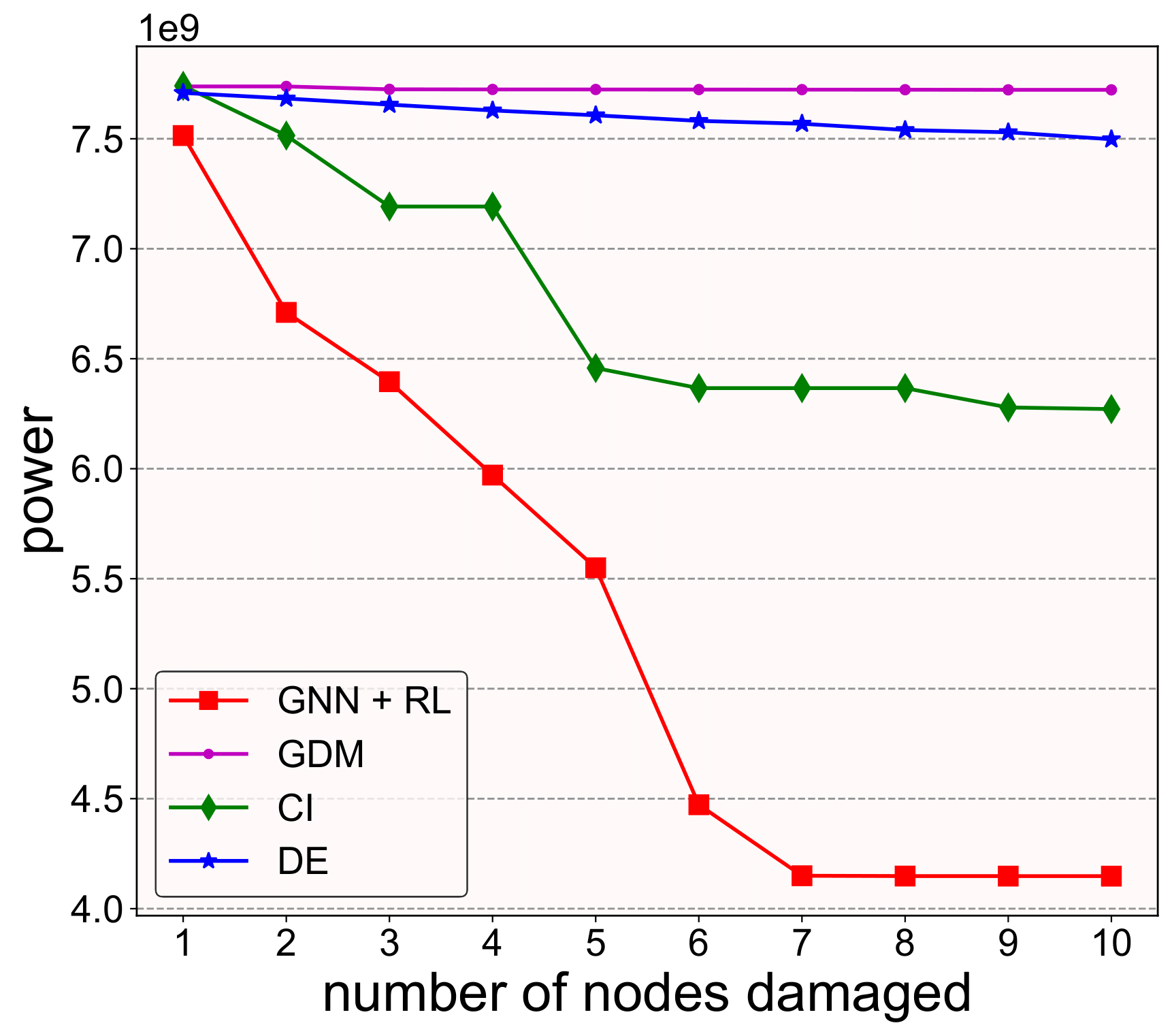}}
  \subfloat[ANC Decrease of Primary Road]{\includegraphics[width=5.5cm]{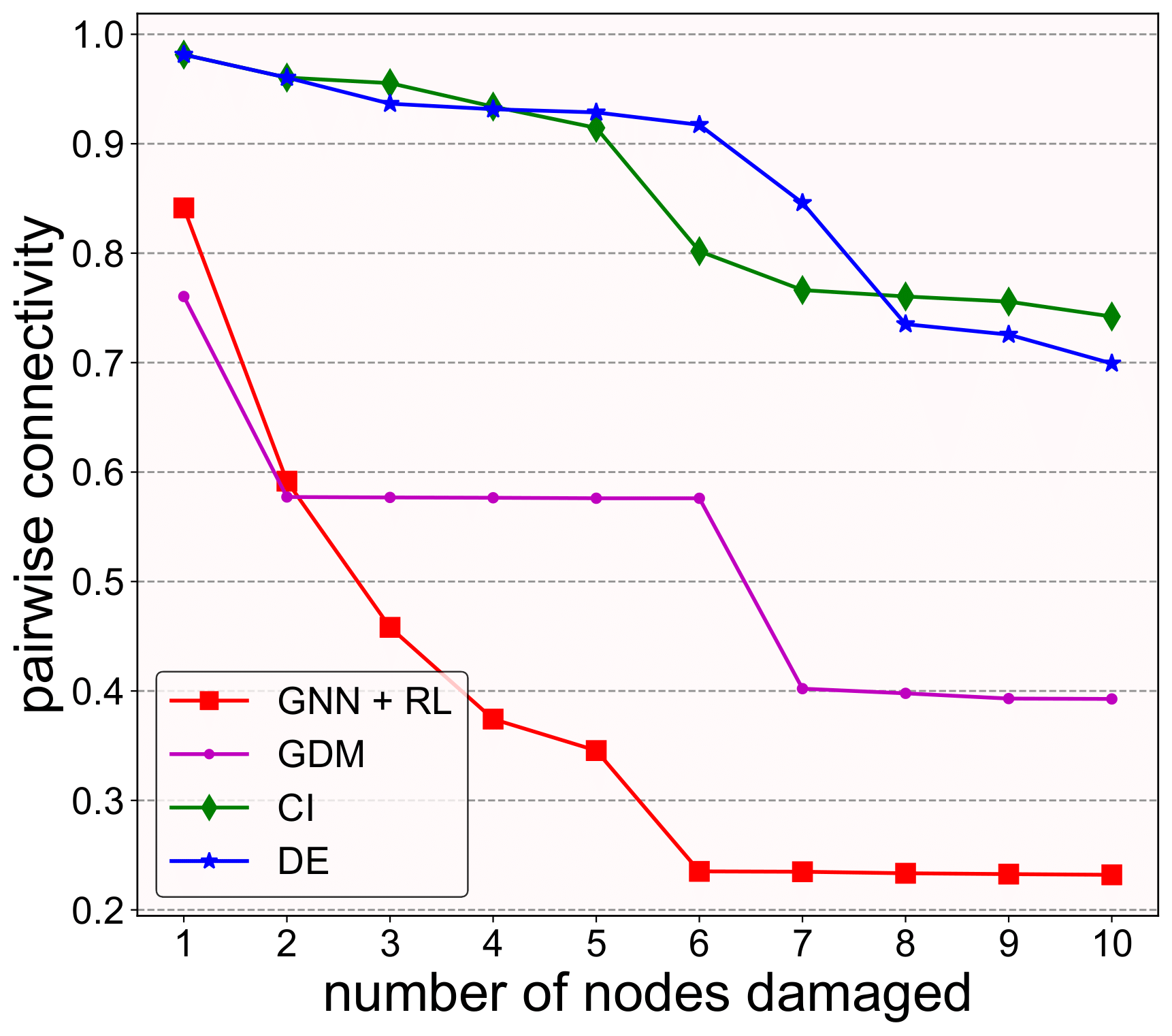}}
    \subfloat[ANC Decrease of Tertiary Road]{\includegraphics[width=5.5cm]{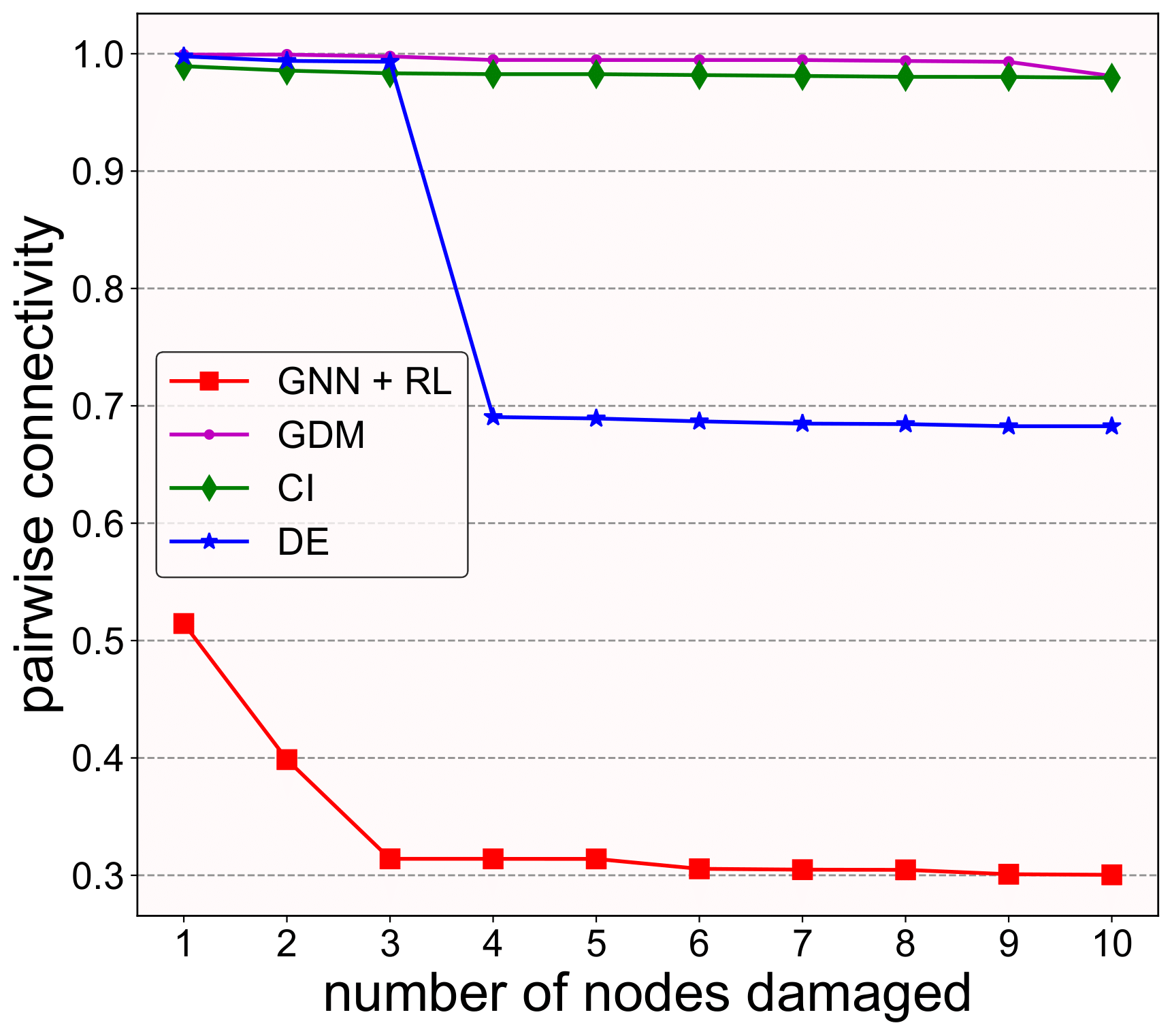}}

    \subfloat[Reward Increase of Bigraph]{\includegraphics[width=5.5cm]{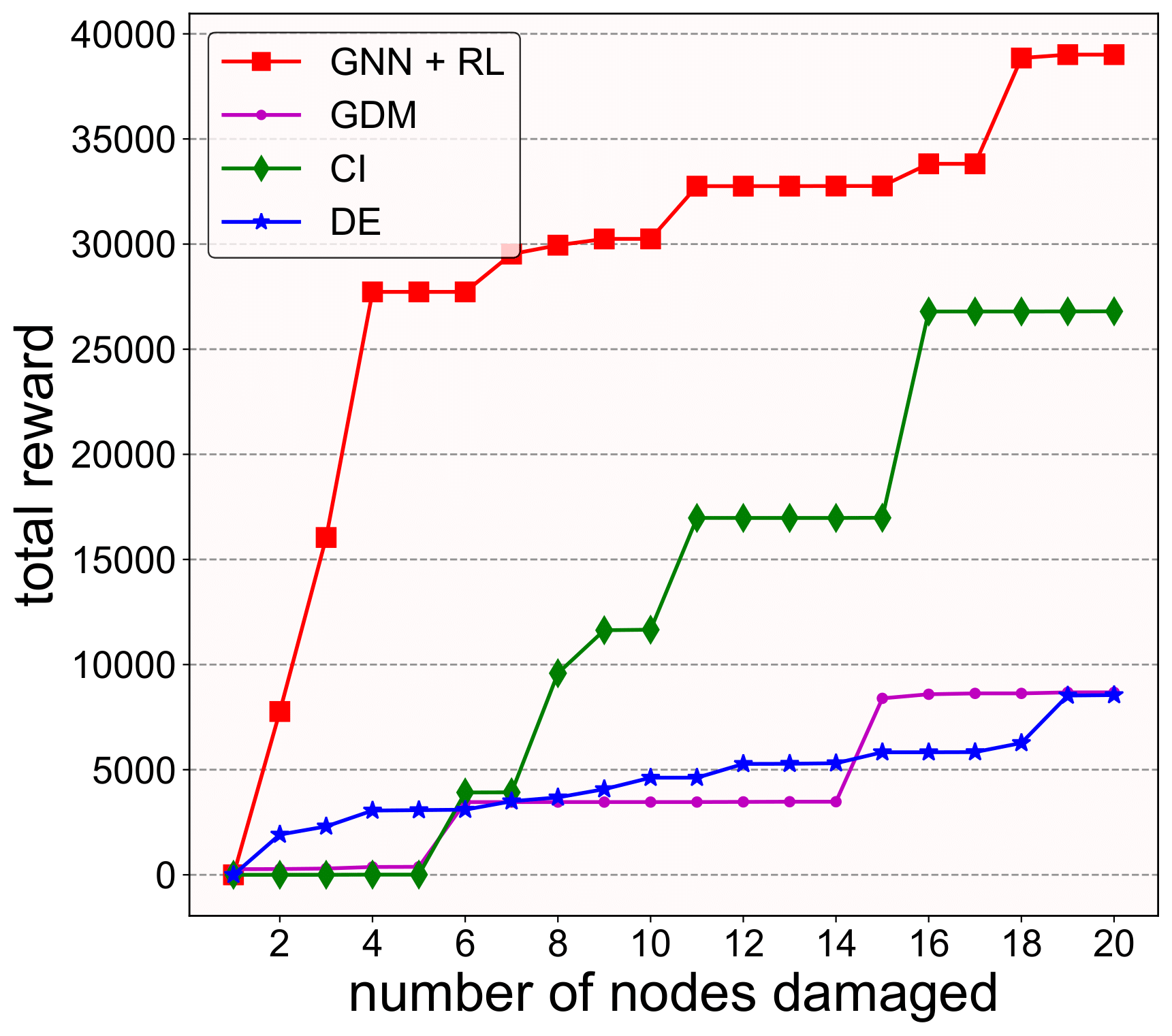}}
    \subfloat[Power Decrease of Bigraph's Electricity ]{\includegraphics[width=5.5cm]{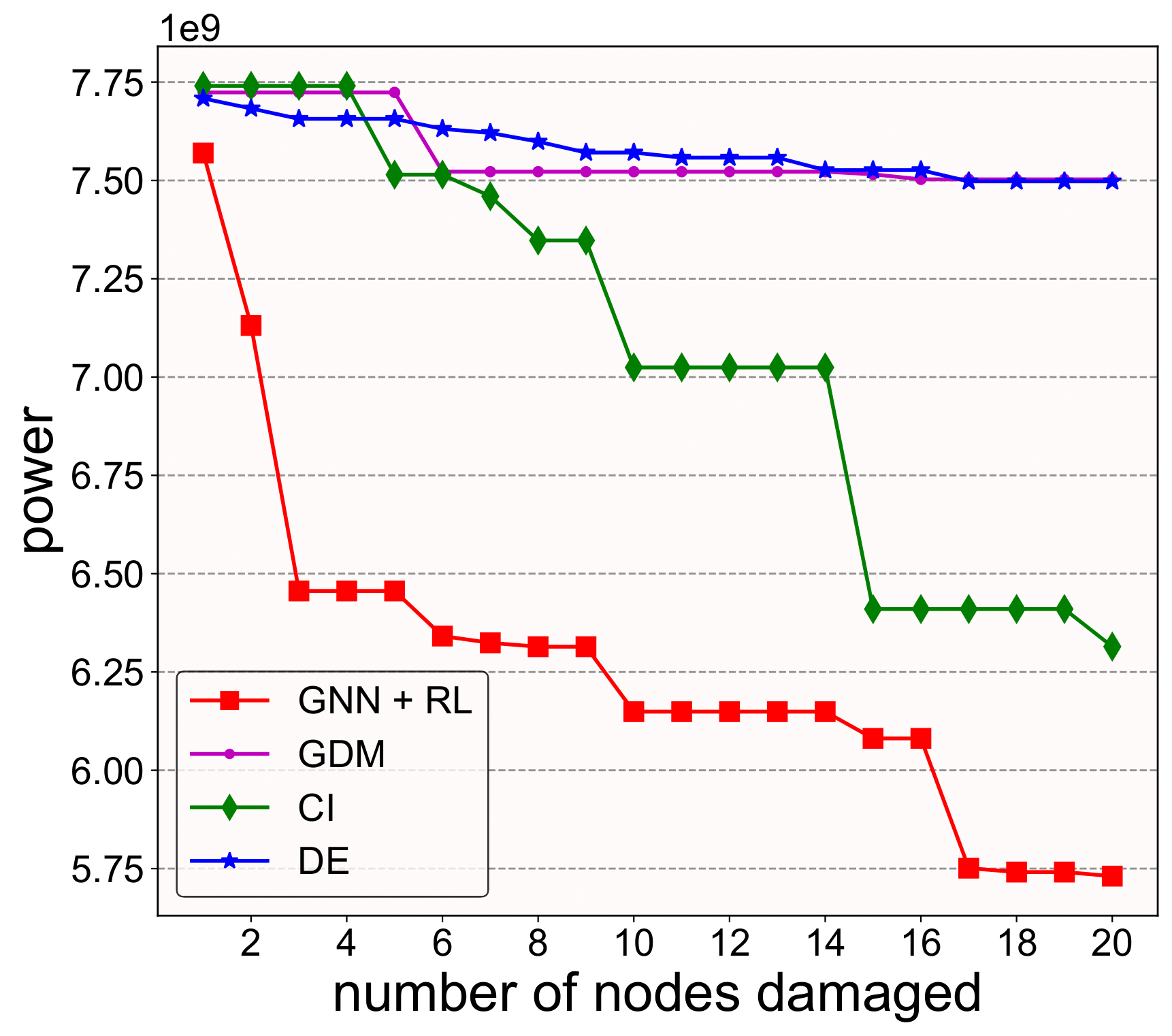}}
    \subfloat[ANC Decrease of Bigraph's Road]{\includegraphics[width=5.5cm]{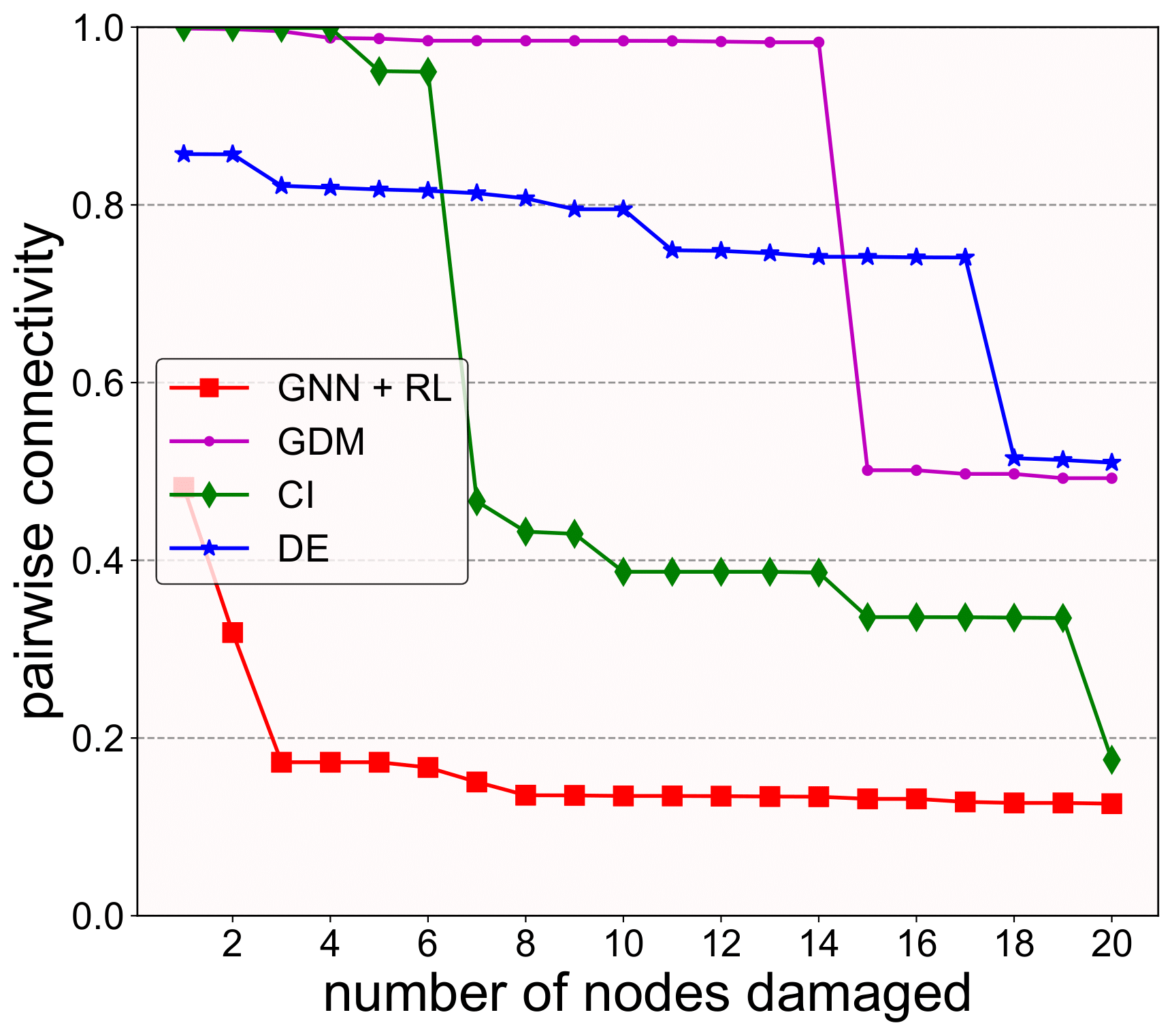}}
  \setlength{\abovecaptionskip}{0.2cm}
        \setlength{\belowcaptionskip}{0.2cm}
  \caption{Performance of different methods on different networks.} 
  \label{fig::main}
\end{figure*}

 In the course of detecting the vulnerable nodes in the coupled graph, based on the strong representation ability of graph neural network (GNN) for graph-structured data and the ability of deep reinforcement learning (DRL) to solve sequential decision-making problem, we devise a new system to deal with this task.
Figure 1 illustrates the architecture of our proposed system.

\subsection{Coupled Graph Representation}
We expect to take into account the direct and indirect information of the graph while choosing nodes, but the state of the graph can be very complex.
Traditional methods\cite{yehezkel2012degree, li2017clustering} usually use fixed features to represent graphs, such as degree distributions, clustering coefficients, and so on. Due to the complex evolutionary mechanism of the coupled graph, these methods are not applicable to most coupled graphs and perform poorly when they characterize the graph's structural information.
Based on graph neural networks, we leverage semi-supervised embedding learning \cite{zhu2005semi} to obtain the coupled graph embedding that could extract the node features and graph topology in a low-dimensional embedding space.
\subsubsection{Graph Construction}
 Given an urban interdependent network consisting of an electricity network and a road network, we represent it as a coupled graph $\mathcal{G}=(\mathcal{V}, \mathcal{E})$ according to the previous setting.
We map each node $v \in \mathcal{V}$ to a initial d-dimensional vectors $\mathbf{f}_v$ and define a node embedding matrix $\mathbf{F}_{v} = \left\{\mathbf{f}_v, \forall v \in \mathcal{V}\right\}\in \mathbb{R}^{d \times |\mathcal{V}|}$ as,
\begin{equation}
\mathbf{f}_{v}=\left\{\begin{aligned}
\mathbf{f}_{v}^e, & \text { if node } {v \in \mathcal{V}^e} \text {;} \\
\mathbf{f}_{v}^r, & \text { if node } {v \in \mathcal{V}^r} \text {.}
\end{aligned}\right.
\end{equation}
Note that the initial embeddings $\mathbf{f}_{v}^e$ and $\mathbf{f}_{v}^r$ are pre-trained from the respective graphs by the method described below, while the initial embeddings of them are randomly set. 
To improve its characterization ability, we use graph neural networks\cite{gao2023survey} to incorporate node attributes and capture the coupled graph structural information and other unseen data. 
On account of initial embeddings that have already contained the information of subgraphs, we are able to explore more information about the interdependent relationship in the coupled graph.
\subsubsection{Graph Neural Network}
Graph $\mathcal{G}=(\mathcal{V}, \mathcal{E})$ and initial embeddings of all nodes $\left\{\mathbf{f}_{v}, \forall v \in \mathcal{V}\right\}$ are provided as input. 
Let $l$ denote the depth and $\mathbf{h}_v^l \in \mathbb{R}^{d \times 1}$ denotes the embedding vector for node $v$ at step $l$, each node $v \in \mathcal{V}$ aggregates the embeddings of its immediate neighbor nodes, $\left\{\mathbf{h}_u^{l}, \forall u \in \mathcal{N}(v)\right\}$, into a single vector $\mathbf{h}_{\mathcal{N}(v)}^{l} \in \mathbb{R}^{d \times 1}$, where $\mathcal{N}(\cdot)$ stands for the set of neighboring nodes.
Note that $\mathbf{h}_v^0$ is defined as the input features $\mathbf{f}_{v}$. The aggregation process is shown as follows,
\begin{equation}
\mathbf{h}_{\mathcal{N}(v)}^{l} = \operatorname{AGG}(\left\{\mathbf{h}_u^{l}, \forall u \in \mathcal{N}(v)\right\}) \text {,}
\end{equation}

\begin{equation}
\mathbf{h}_v^{l+1} = \sigma\left(\mathbf{W}^l \cdot \operatorname{MEAN}(\left\{\mathbf{h}_v^{l},\mathbf{h}_{\mathcal{N}(v)}^{l}\right\})\right) \text {,}
\end{equation}
where $\operatorname{AGG}$ is a summation function that aggregates information from local neighbors \cite{kipf2016semi}, $\sigma$ stands for the rectifier function  $\operatorname{ReLU}$, $\operatorname{MEAN}$ is the mean function, and $\mathbf{W}_1^l \in \mathbb{R}^{d \times d}$ is a weight matrix to be learned at step $l+1$. 
By defining the depth or iterations $L$, we can get the final representation for all nodes at depth $L$, $ \left\{\mathbf{z}_v \equiv \mathbf{h}_v^L, \forall v \in \mathcal{V}\right\}$, which contains $L$-order neighbor nodes' information. 
\subsubsection{Semi-supervised Embedding Learning}

There are cascade relationships between infrastructures that are not connected directly in the urban network,
and the topology of a network implicitly reveal the signal of similarity and correlation among nodes, which can be used as supervised signals to obtain the representation vectors. Therefore, the task can be treated as  a link prediction \cite{grover2016node2vec} problem, which aims to estimate the likelihood of the edges between nodes, based on observed edges and attributes of nodes. Then we take all of the edges in the graph as a positive set and sample several edges that do not exist as a negative set, which forms a positive graph $\mathcal{G}^p$ and a negative graph $\mathcal{G}^n$, respectively. 
To learn parameters of the aggregator function and weight matrices $\left\{\mathbf{W}^l, \forall l \in  \left\{1,\ldots, L\right\}\right\}$ via stochastic gradient descent, we firstly define a link prediction function as follows,
\begin{equation}
\operatorname{S}(\mathbf{Z}_v,\mathcal{G}) = \left\{\mathbf{z}_v^T \cdot \mathbf{z}_u, \forall e_{vu} \in \mathcal{E}  \right\} \text {,}
\end{equation}
which calculates the inner product of embedding vectors for two nodes on each existing edge as predicted weight, on given graph $\mathcal{G}$. Then we apply a margin loss function \cite{rosset2003margin} to the node embeddings $\mathbf{Z}_v$ as follows, 
\begin{equation}
\small
J_{\mathcal{G}}\left(\mathbf{Z}_v\right)=\operatorname{MEAN}(\max (0, M - 
\operatorname{S}(\mathbf{Z}_v,\mathcal{G}^p) + \operatorname{S}(\mathbf{Z}_v,\mathcal{G}^n)) + \lambda\ \Vert \Theta\ \Vert ^2 \text {,}
\end{equation}
where $M$ is a constant parameter, the second term of the loss function performs $L2$ regularization where $\Theta$ stands for model parameters and $\lambda$ controls the penalty strength.
For $\mathcal{G}^e$ and $\mathcal{G}^p$, the weights simply represent the presence or absence of edges, and for coupled graph, $\mathcal{G}$, weights are set to even indicate the type of the edges.
While unknown edges are predicted by the known edges, we can obtain more superior embeddings for the interdependent network with cascading relationships, through semi-supervised learning \cite{wang2019dgl}.

\begin{figure*}[htbp]
  \centering
  \subfloat[Transferring Ability for Electricity Network]{\includegraphics[width=5.5cm]{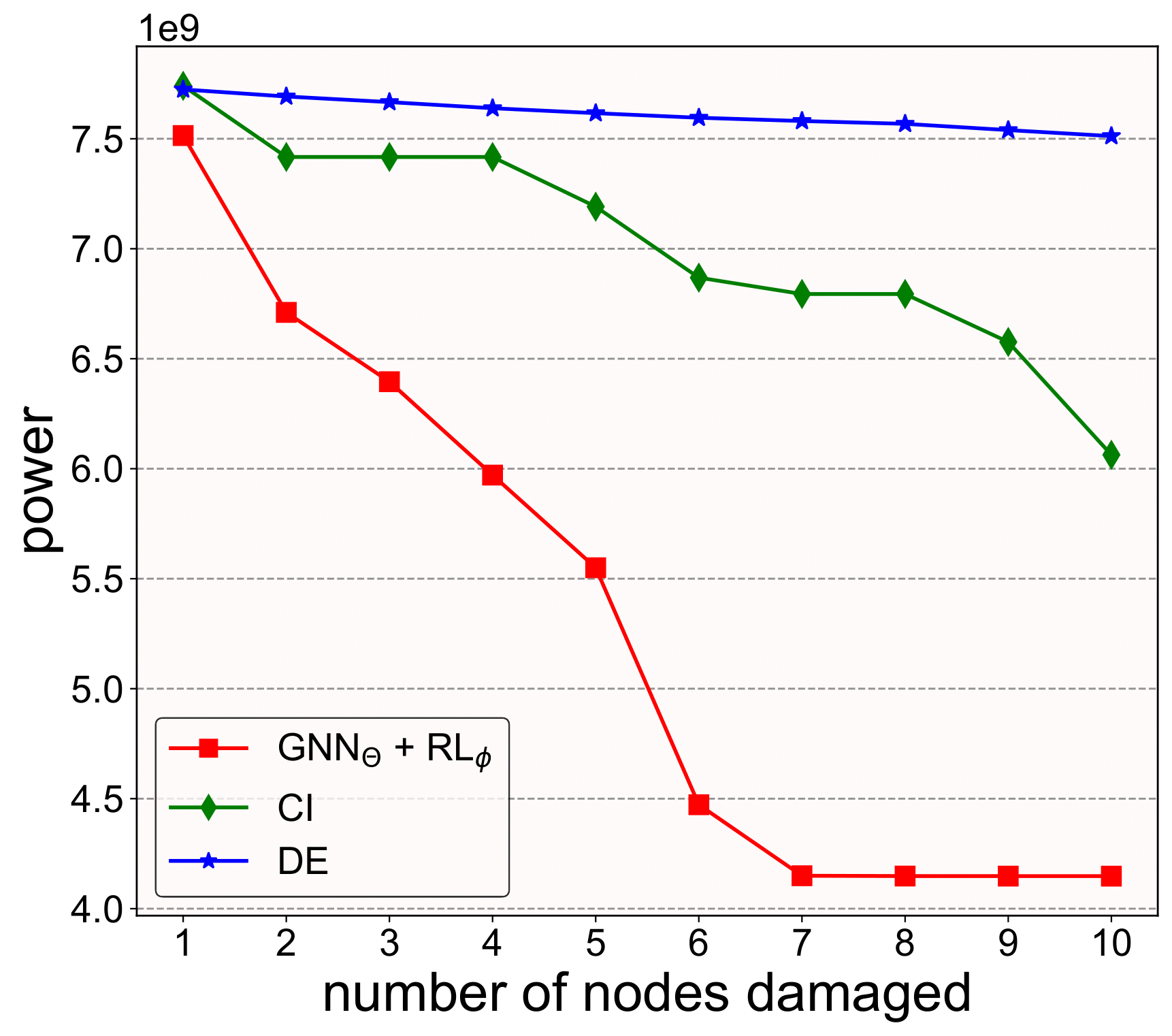}}
  \subfloat[Transferring Ability for Primary Road]{\includegraphics[width=5.5cm]{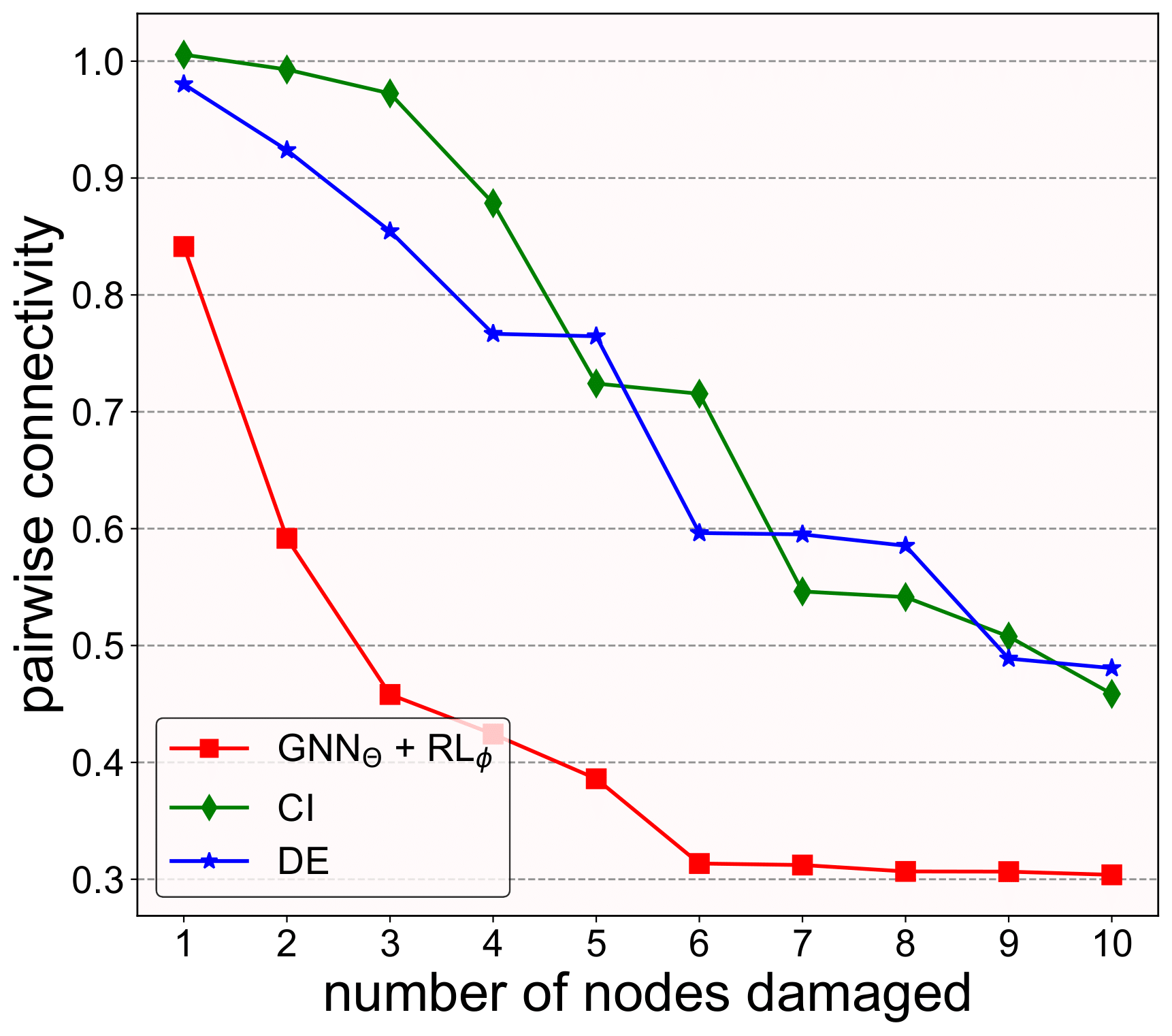}}
    \subfloat[Transferring Ability for Coupled Network]{\includegraphics[width=5.5cm]{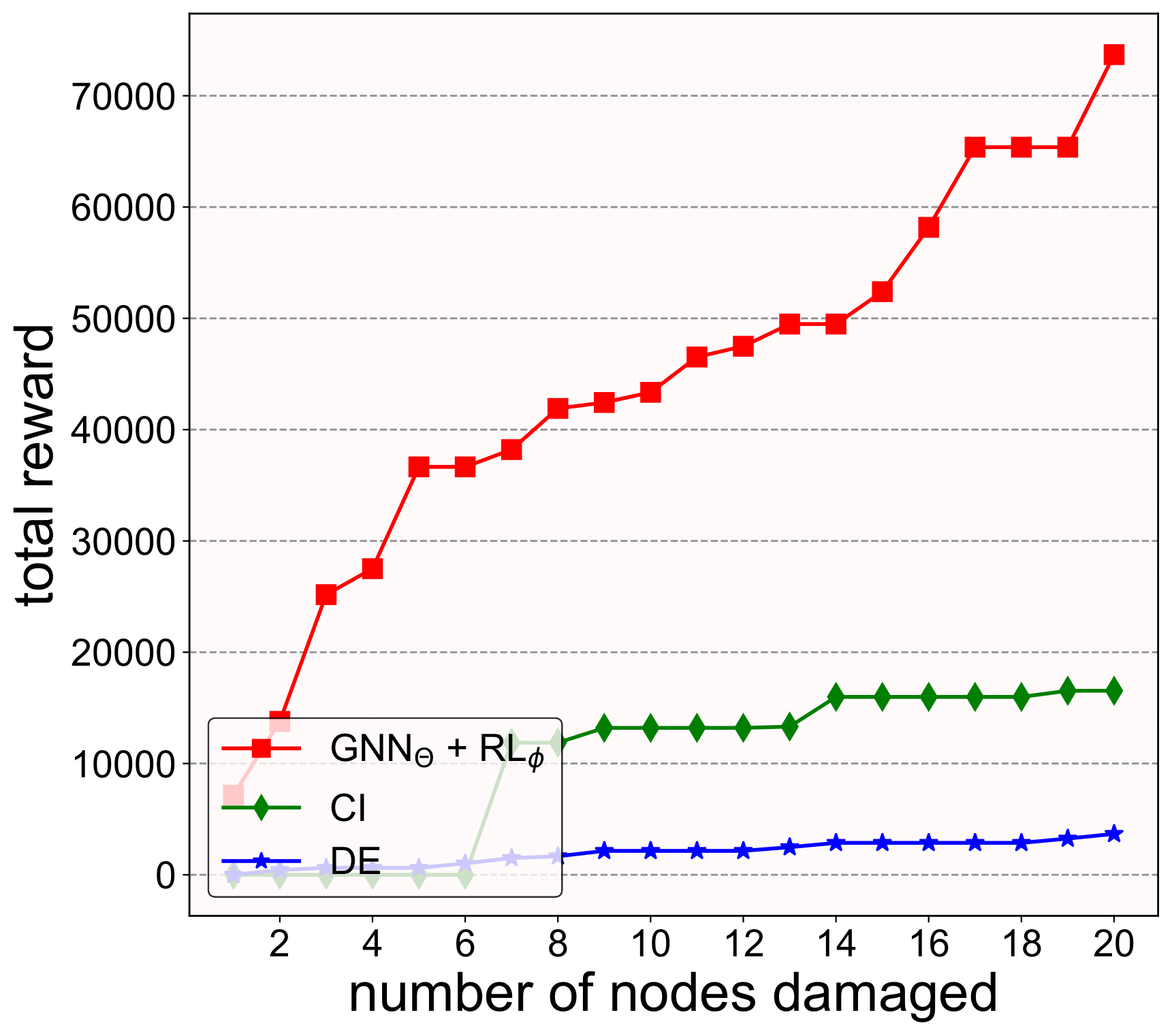}}

  \setlength{\abovecaptionskip}{0.2cm}
  \setlength{\belowcaptionskip}{0.2cm}
  \caption{Transferring ability for different Networks.} 
  \label{fig::transfer}
\end{figure*}

\subsection{Vulnerable Nodes Detecting}
We obtain a coupled graph embedding matrix through a graph convolutional neural network, which can efficiently extract the topology features of the interdependent network. 
To make full use of this information, and find a set of vulnerable nodes in the network, we construct a reinforcement learning module. 
Reinforcement learning \cite{sutton2018reinforcement} is a sequential decision process where the agent is trained to take optimal actions for different scenarios of an environment. 
We formulate the RL problem as a Markov decision process (MDP), which is defined by a tuple $\left(\mathcal{S}, \mathcal{A}, \mathcal{P}, \mathcal{R}, \gamma\right)$ with a set of states $\mathcal{S}$, a set of actions $\mathcal{A}$, a reward function $\mathcal{R}\left(s, a\right)$, the transition function $\mathcal{P}(s^{\prime}\mid s, a)$ and a discount factor $\gamma$, $s, s^{\prime} \in \mathcal{S}, a \in \mathcal{A}$. $\mathcal{P}(s^{\prime}\mid s, a)$ governs the joint probability distribution $p\left(s^{\prime}\mid s, a\right)$ of transitioning to new state $s^{\prime}$ after taking action $a$ in state $s$. 
Given a policy $\pi$ , which fully determines the agent' behavior, the action-value function $Q_\pi(s, a)$ is defined as the expected reward when starting from $s$, taking action $a$, and following the policy $\pi$. In our proposed system, the action is to choose vulnerable nodes, the state is the residual coupled graph being analyzed after chosen nodes are damaged and the reward is the resulting weighted summation of cascade failure from the environment which represents the given interdependent network.
\subsubsection{Framework}
Given the couple graph $\mathcal{G}=(\mathcal{V}, \mathcal{E})$ and the corresponding pre-trained node embeddings $\mathbf{Z}_v \in \mathbb{R}^{d \times |\mathcal{V}|}$ as the input of our model, we expect to sequentially select $K$ vulnerable nodes from the graph as output for a complete epoch.
 We use the node embedding matrix $\mathbf{Z}_v$ to represent the universal state $s$ of the current network (environment), and the initial state can be defined as the graph pooling, $\mathbf{s}_1 = \operatorname{MEAN}(\mathbf{Z}_v) \in \mathbb{R}^{d \times 1}$. 
Then at each step $k$, the agent selects a node $v_k \sim \pi(\cdot \mid \mathbf{s}_k)$ as the action, where $\pi$ is the policy. At the same time, next state $\mathbf{s}_{k+1}$ is calculated by removing node $v_k$ as follows,
\begin{equation}
\mathbf{s}_{k+1} = \operatorname{MEAN}(\mathbf{Z}_v\backslash\left\{v_1,\ldots,v_k\right\}) \text {,}
\end{equation}
where $\mathbf{Z}_v\backslash\left\{v_1,\ldots,v_k,\right\}$ is the residual node embeddings after removing embeddings of nodes in the set $\left\{v_1,\ldots,v_k\right\} \subseteq  \mathcal{V}$. 
Then the environment judges whether selected node $v_k$ is a power station or a junction, and set it damaged in the urban network to obtain the corresponding reward $r_k$.
The agent update its policy $\pi(v_k \mid \mathbf{s}_k)$ with reward $r_k$, and choose the next node $v_{k+1}$. 
Continue this process until epoch ends, and a trajectory $\tau=\left(\mathbf{s}_1, v_1, r_1, \ldots,\mathbf{s}_K, v_K, r_K\right)$ is determined by the agent under the policy $\pi$. 
The learning objective is to discover an optimal policy $\pi^*$ that can maximize the total reward for this whole trajectory: $\sum_{k \geq 1} \gamma^k r_k$. Based on \emph{Bellman equation} \cite{sutton2018reinforcement}, the optimal action-value function $Q_{\pi^*}(\mathbf{s}, v)$ can be derived as:
\begin{equation}
Q_{\pi^*}(\mathbf{s}, v)=\sum_{\mathbf{s}^{\prime} \in \mathcal{S}} p\left(\mathbf{s}^{\prime} \mid \mathbf{s}, v\right)\left[\mathcal{R}\left(v\right)+\gamma \max _{v^{\prime}} Q^*\left(\mathbf{s}^{\prime}, v^{\prime}\right)\right] \text {,}
\end{equation}
A common method of obtaining a new policy for an action-value function is to adopt $\epsilon$-greedy policy \cite{hessel2018rainbow}, which means that the agent either chooses the node with the highest value with probability $1-\epsilon$ or chooses a node randomly with probability $\epsilon$. It has been proved that the agent following $\epsilon$-greedy policy could obtain the maximum total rewards when  $Q_{\pi}(s, v)$ is optimal \cite{szepesvari2010algorithms}.
\subsubsection{Value Network and Optimization}

Owing to the agent's action that choosing a node at each step is discrete, and the action space is extremely large we design our vulnerable nodes detecting model inspired by Deep Q-Learning Network (DQN) algorithm \cite{mnih2015human}, which takes advantage of a convolution neural network to approximate the $Q_{\pi}(\mathbf{s}, v)$. 
To make full use of the convolution neural network and structural information learned by the node embeddings $\mathbf{Z}_v$ when choosing nodes, the action-value function $Q_{\pi}(\mathbf{s}, v)$ are defined as follows,
\begin{equation}
Q_{\pi}(\mathbf{s}_k, v_k;\Theta) = \operatorname{MEAN}((\mathbf{\theta}_2\cdot(\sigma(\mathbf{\theta}_1 \cdot \mathbf{Z}_v)))^T \cdot \left[ \mathbf{s}_k, \ldots, \mathbf{s}_k \right]_{|\mathcal{V}|})\text{,}
\end{equation}
where $\mathbf{\theta}_1 \in \mathbb{R}^{2d \times d}$ and $\mathbf{\theta}_2 \in \mathbb{R}^{d \times 2d}$ are the optimized parameters, $\sigma$ stands for a rectifier function and $\left[ \mathbf{s}_k, \ldots, \mathbf{s}_k \right]_{|\mathcal{V}|}$ is a function that concatenating $\mathbf{s}_k$  for $\mid \mathcal{V} \mid$ times. 
At each step, $Q_{\pi}(\mathbf{s}_k, v_{k-1})$ is used for the $\epsilon$-greedy selection of the next action $v_k$m, and the agent stores the transition tuple $(\mathbf{s}_k, v_k,r_k,\mathbf{s}_{k+1})$ into an experience replay buffer, the size of which is $E$. 
The parameters $\Theta$ are trained over tuples sampled uniformly from the reply buffer via stochastic gradient to minimize the mean-squared loss descent  :
\begin{equation}
\small
J_Q(\Theta) = \operatorname{MEAN}(\Vert (r_k+ 
\gamma \max_v \hat{Q}(\mathbf{s}_{k+1}, v ; \hat{\Theta})) - Q(\mathbf{s}_k, v_k ; \Theta) \Vert^2) \text{,}
\end{equation}
where $\hat{\Theta}$ represent the parameters of the target net $\hat{Q}$, which only updates after several iterations by copying $\Theta$ of evaluation net $Q$. This approach ensures a more stable and efficient training process.
Since the graph is extracted from real data, we obtain the reward $r$ by simulating the network changes after chosen nodes are damaged.
To detect the vulnerable infrastructures in the road network, given the chosen node $v_k$, reward $r$ is the weighted summation of the decreased power of the electric network and decreased value of connectivity of the road network. 
At step $k$,  the reward function $\mathcal{R}(v_k)$ are defined as follows, 
\begin{equation}
\mathcal{R}(v_k) = \left\{\begin{aligned}
a^eP(v_k, \mathcal{G}^e)& +  a^rA(E(v_k, \mathcal{G}), \mathcal{G}^e),  &\text { if node } {v_k \in \mathcal{V}^e} \text {;}  \\
&a^rA(v_k, \mathcal{G}^e),  &\text { if node } {v_k \in \mathcal{V}^r} \text {;}
\end{aligned}\right.
\end{equation}
where $a^e$ and $a^r$ are weight coefficients, $P(v_k, \mathcal{G}^e)$ is a function that can obtain the decreased power of given electricity network $\mathcal{G}^e$ after power station $v^e_k$ is damaged by Power Flow Calculation,  $E(v_k, \mathcal{G})$ is a function that returns a set of invalid traffic lights affected by damaged power station $v^e_k$ and $A(v_k, \mathcal{G}^e)$ is a function that can calculate the decreased value of connectivity of the road network. We can set a reasonable reward by adjusting the weight coefficients, to detect the ideal vulnerable nodes on the coupled graph.

\section{Evaluation}\label{sec::experiments}

In this section, we aim to evaluate the proposed model with extensive  answers to the following research questions (RQs).
\begin{itemize}[leftmargin=*]
    \item RQ1: Could our proposed method achieve the best performance compared with existing methods?
    \item RQ2: What about the transferability of our model? Specifically, how about the model's performance when transferred into a new interdependent network?
    \item RQ3: What about the effectiveness of each component in the proposed method?
    \item RQ4: Why can our method find the vulnerable nodes in the graph? What about the failure process in the interdependent network?
\end{itemize}

\subsection{Experimental Settings}

\subsubsection{Experimental environment}
For the single network, we leverage the electricity network $\mathcal{G}^{e}$ with 10,887 nodes and 11,438 edges, including different levels of power stations such as 550kV and 220kV, and infrastructure such as traffic lights. In addition, two different road networks---primary road $\mathcal{G}^{rp}$ with 1,035 nodes and 1,161 edges, and tertiary road $\mathcal{G}^{rt}$ with 4,825 nodes and 5,025 edges included, are also applied the model introduced before to choose vulnerable nodes, using metrics followed in Section 4.1.2. 
For the interdependent network with 15,712 nodes and 21,191 edges,
representing the coupled electricity network and the tertiary road we attack nodes consisting of two graphs, importing the basic rules that once a power station fails, the corresponding traffic lights stop working.

\subsubsection{Metrics} \label{metrics}

We use the representative metrics to evaluate the selection performance of the proposed model and the existing method as follows.
\begin{itemize}[leftmargin=*]
    \item \textbf{Electricity Network:} To assess the impact on the electricity network, we calculate the current power after each node is damaged or disabled using flow calculation within the electricity network.
    \item \textbf{Road Network:} For the road network, we use ratio of connectivity \cite{fan2020finding}, which is defined as follows,
\begin{equation}
\sigma(\mathcal{G}) = \sum_{C_i \in \mathcal{G}} \frac{\delta_i(\delta_i -1)}{2},
\end{equation}
where $C_i$ is the $i$th connected component in the current graph $\mathcal{G}$, and $\delta_i$ is the size of $C_i$.
    Another metric for the road network is the size of GCC, which refers to the giant connected component, calculated by:
\begin{equation}
    \sigma_{gcc}(\mathcal{G}) =\operatorname{max}\{\delta_i; C_i \in \mathcal{G}\}.
\end{equation}
    
    \item \textbf{Coupled Network} For the coupled network, we use the reward defined as the weighted summation of power or connectivity decrease after each node is damaged in the electricity network or road topology correspondingly, to jointly evaluate the performance on two networks.
\end{itemize}

\begin{figure*}[t]
  \centering
  \subfloat[Power Decrease of Electricity Network]{\includegraphics[width=5.5cm]{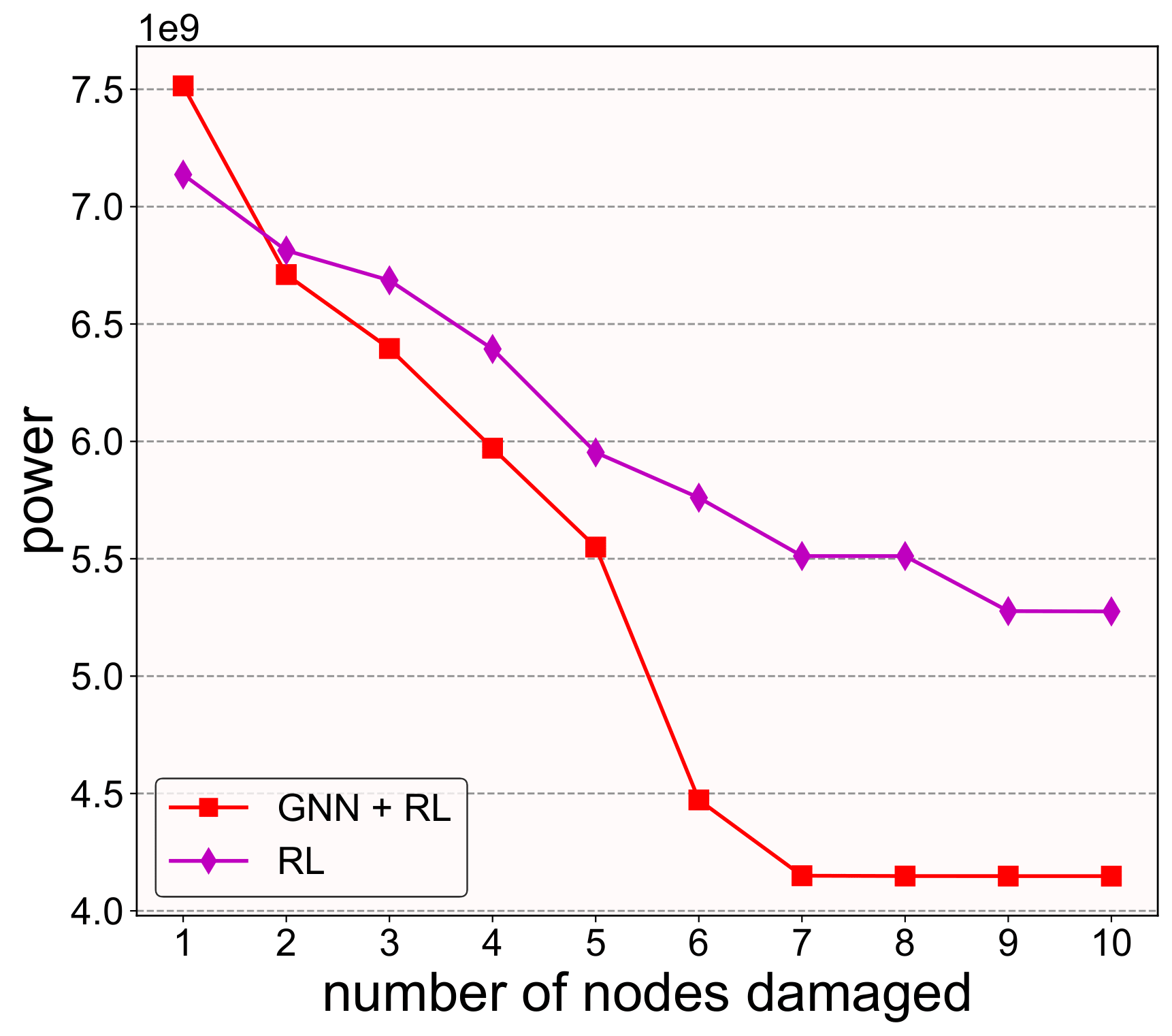}}
  \subfloat[ANC Decrease of Primary Road]{\includegraphics[width=5.5cm]{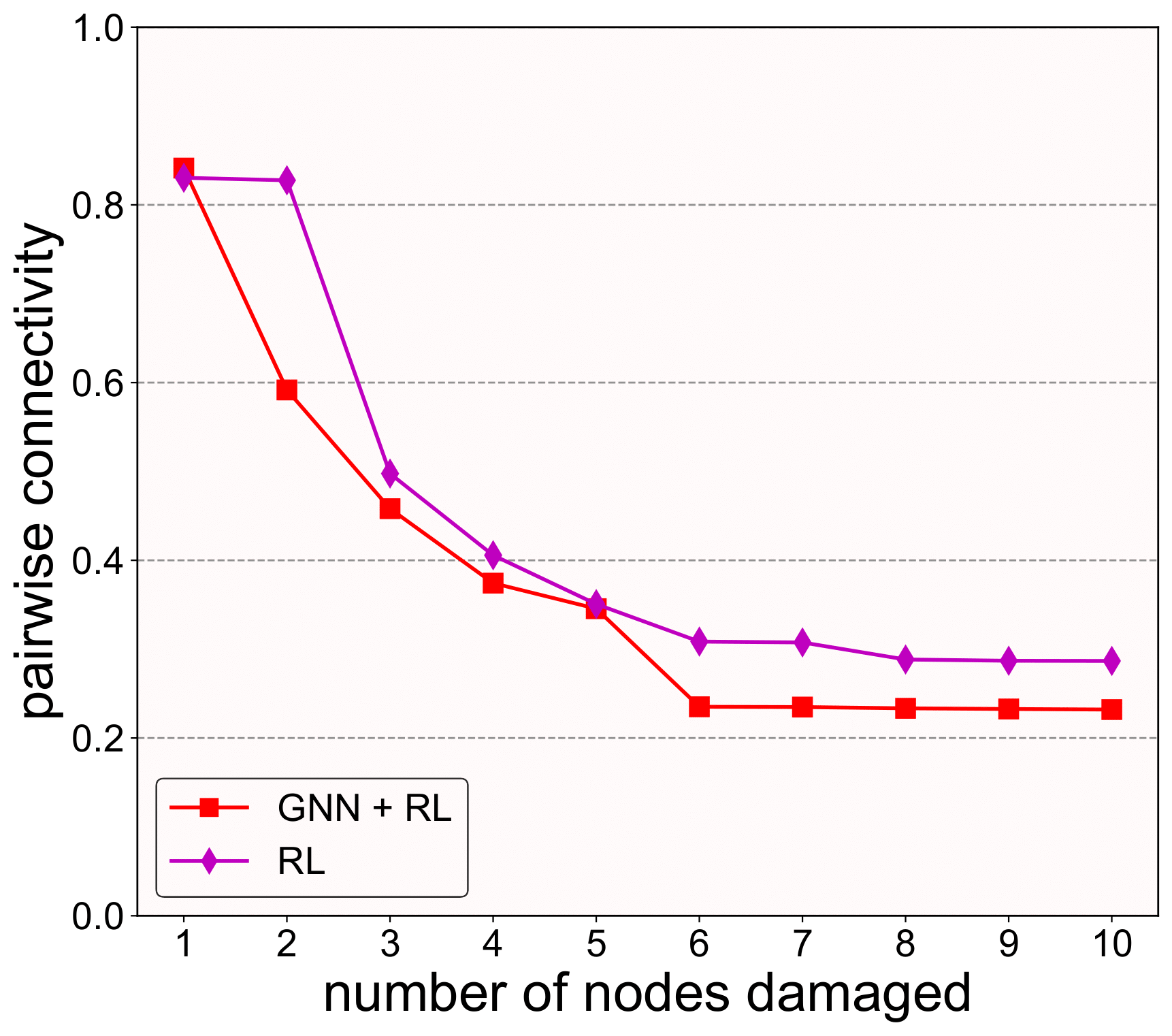}}
  \subfloat[ANC Decrease of Tertiary Road]{\includegraphics[width=5.5cm]{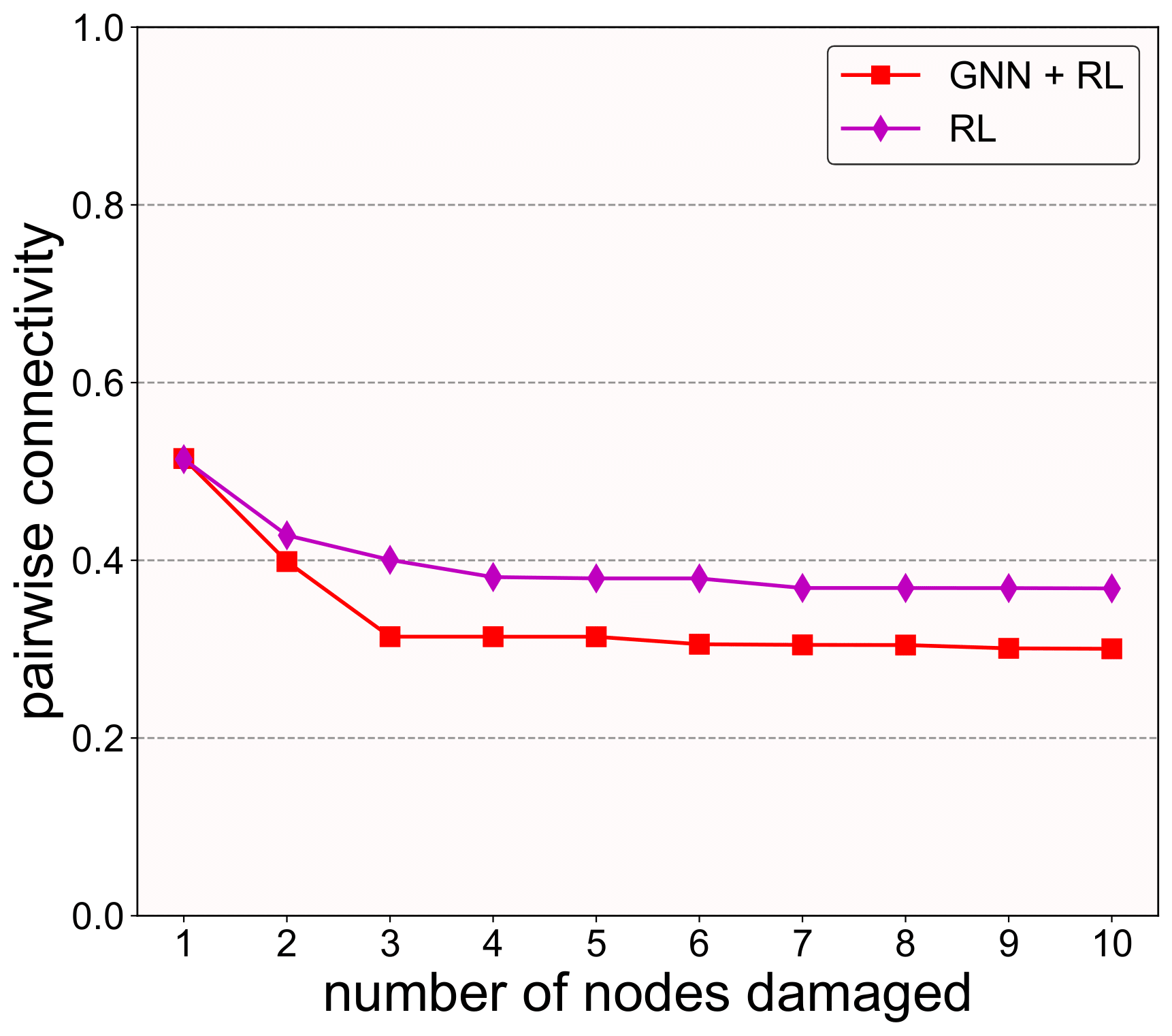}}
  
  \subfloat[Reward Increase for Bigraph]{\includegraphics[width=5.5cm]{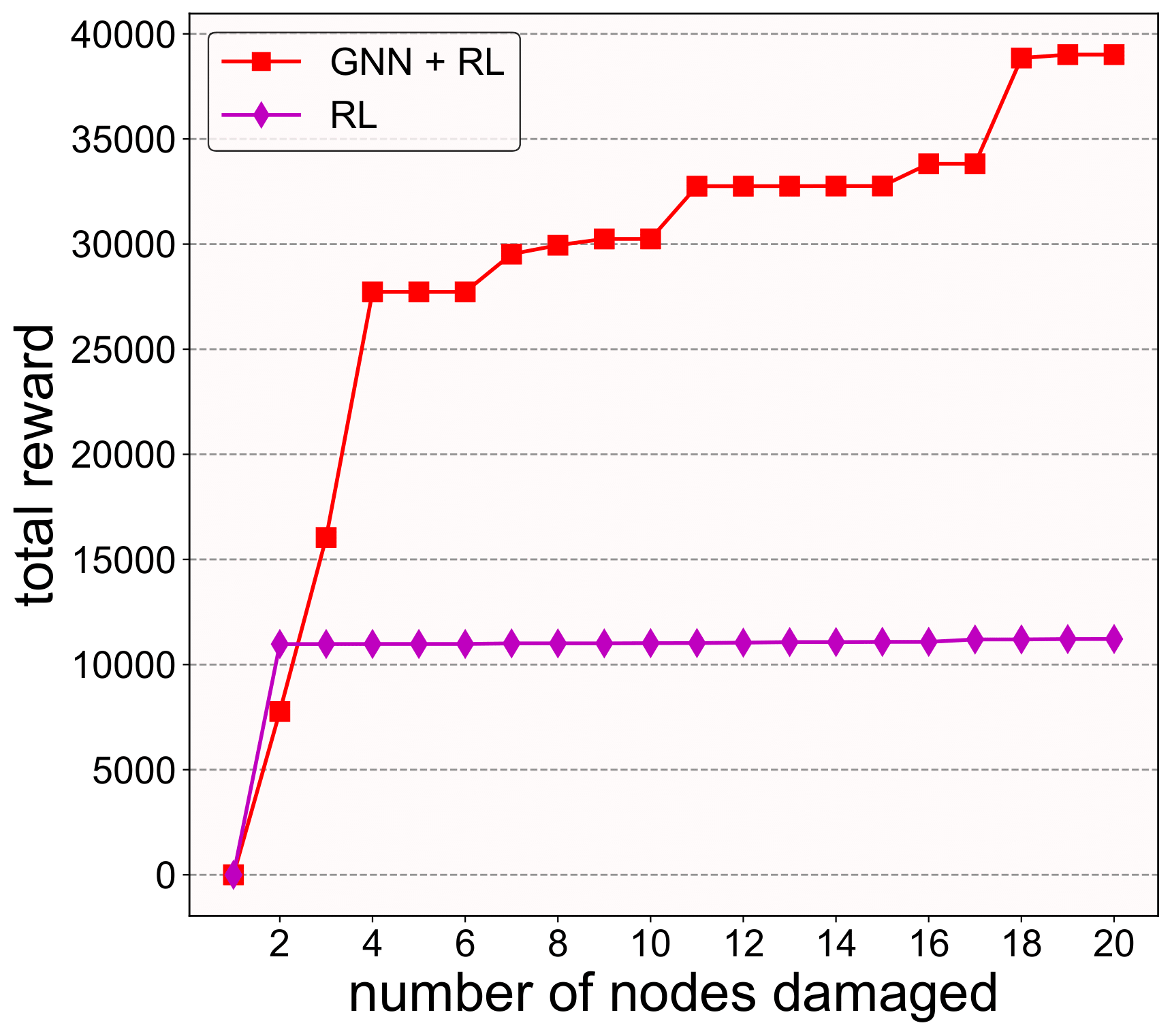}}
  \subfloat[Power Decrease in Bigraph's Electricity]{\includegraphics[width=5.5cm]{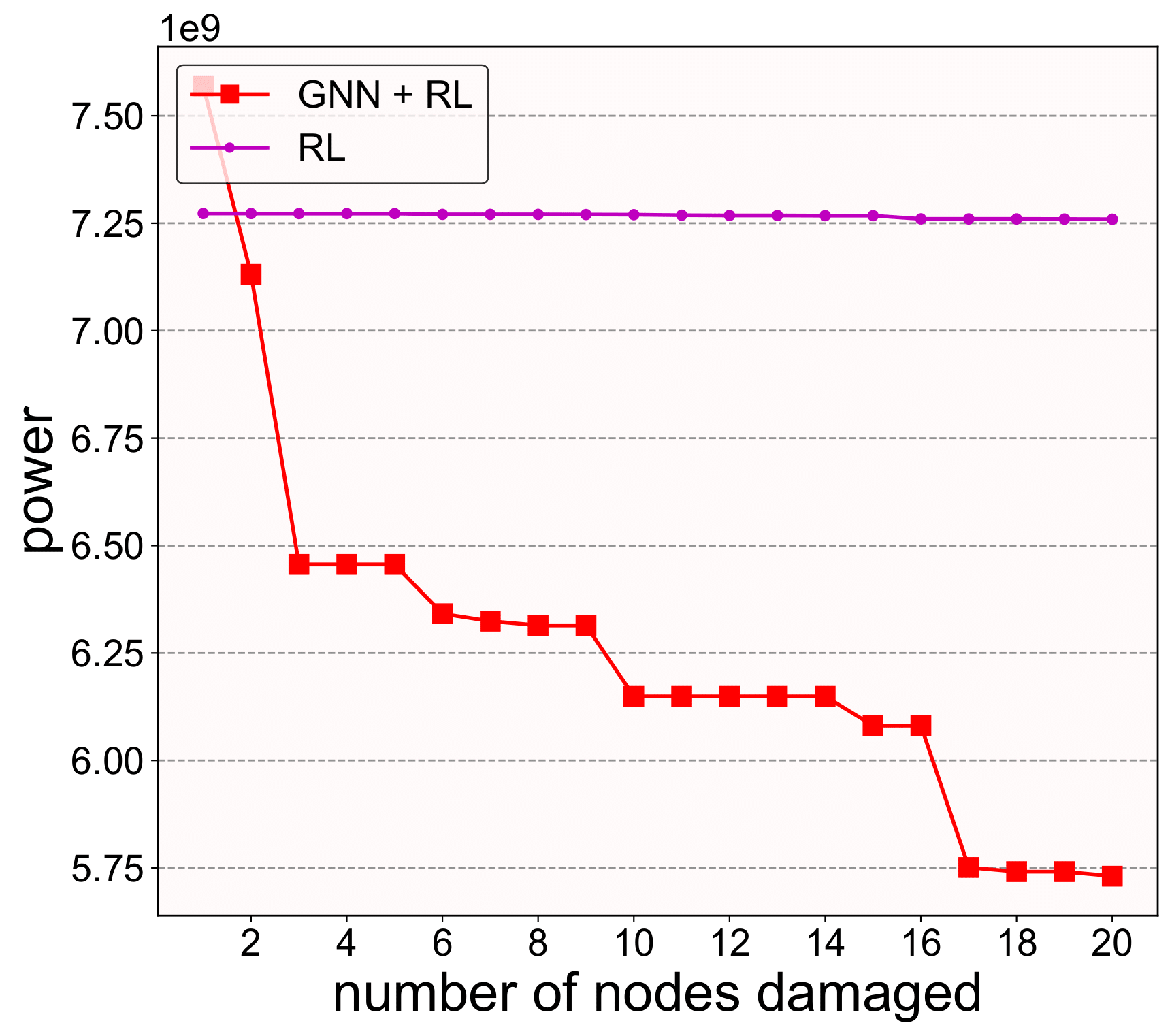}}
  \subfloat[ANC Decrease in Bigraph's Road]{\includegraphics[width=5.5cm]{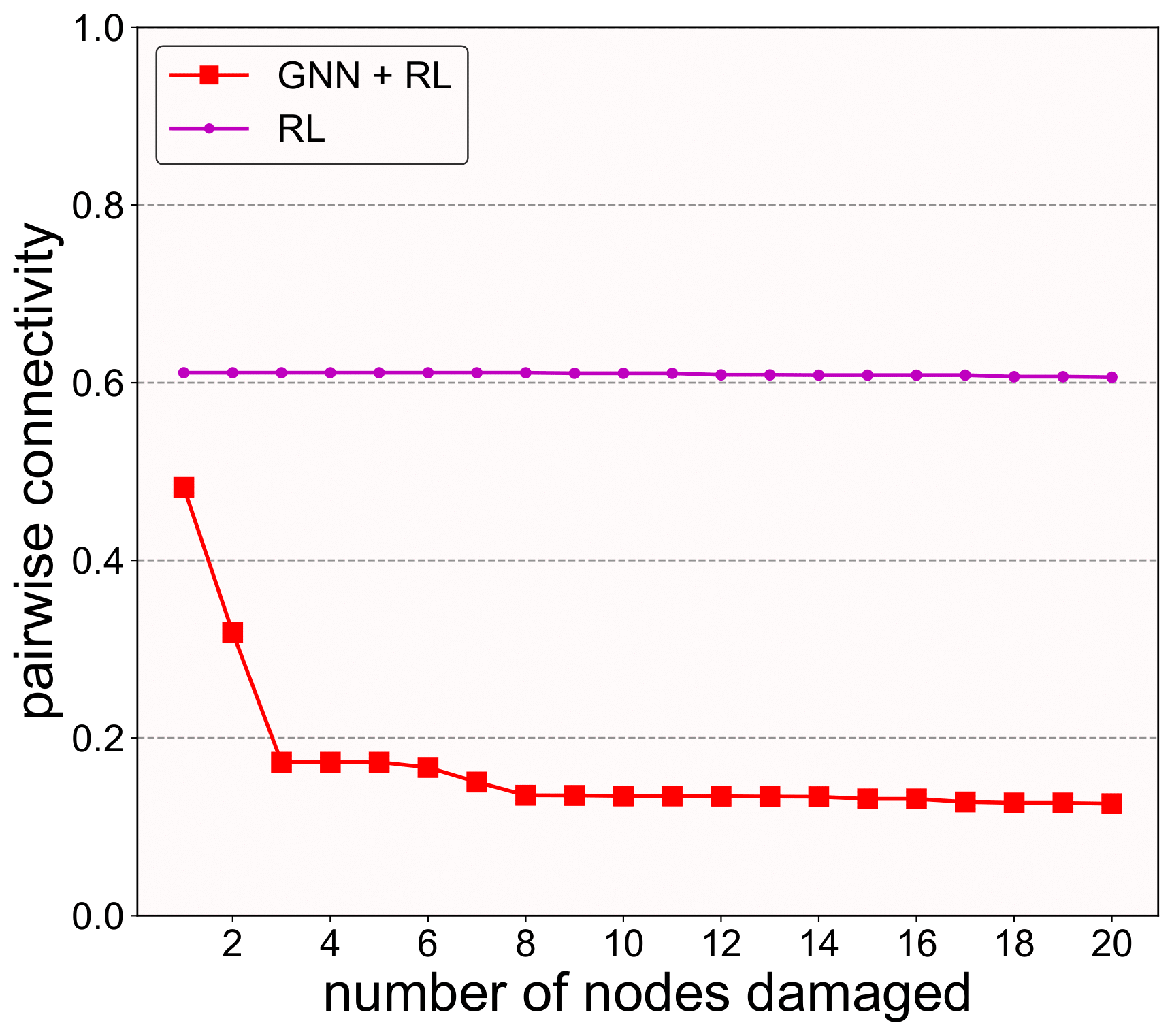}}

  \setlength{\abovecaptionskip}{0.2cm}
        \setlength{\belowcaptionskip}{0.2cm}
  \caption{Ablation experiments for different networks.} 
  \label{fig::abla}
\end{figure*}

\subsubsection{Baselines}
We compare our proposed method with the following competitive baselines.

\begin{itemize}[leftmargin=*]
    \item \textbf{DE~\cite{yehezkel2012degree, iyer2013attack}:} Here we represent the degree of node $v_i$ as $d_i$. This method sorts the degree $d_i$ of each node $v_i$ in the graph and chooses the greatest ten nodes to be removed sequentially.
    \item \textbf{CI~\cite{morone2015influence}:} The Collective Influence (CI) measure is defined as the product of the node’s reduced degree  and the sum of the reduced degrees of the nodes that are within a constant hop away from it. This method is  calculated by $\mathrm{CI}(v)=\left(d_v-1\right) \sum_{u \in \mathcal{N}(v)}\left(d_u-1\right)$, for each node, which describes the proportion of other nodes that can be reached from a given node, assuming the nodes with higher CI values play more crucial roles in networks. The CI method sequentially removes the node with the highest CI value and recalculates the CI values of the remaining nodes for the rest following operations.
    \item \textbf{GDM~\cite{grassia2021machine}:} Graph Dismantling with Machine Learning (GDM), utilizes a supervised learning approach to dismantle the entire network. To generate the training data and labels, we randomly sample nodes from the complete set and classify them into positive and negative sets. We then calculate the ratio of nodes belonging to the positive set for those that exist in both sets. The pre-trained node embeddings obtained from a graph neural network are used as inputs to a neural network, such as a Multilayer Perceptron (MLP), for the supervised learning process. The neural network computes outputs for all nodes, which are subsequently sorted in descending order to identify nodes for removal as a comparison to our method.
\end{itemize}

\subsection{Overall Performance Comparison (RQ1)} \label{RQ1}
We first present the overall performance comparison in Figure \ref{fig::main}. In figure \textbf{(a)-(c)} we demonstrate the testing results on single networks including one electricity network and two road networks. In figure \textbf{(d)-(e)} we demonstrate the testing results on coupled network consisting of electricity network and tertiary road network. Based on the results, we have the following observations.
\begin{itemize}[leftmargin=*]
    \item \textbf{Performance varies by different baselines.}
    Based on observations, we have noticed distinct performance differences among various baselines when applied to single and coupled networks. In the electricity network, CI identifies nodes as more vulnerable compared to DE due to the clustering distribution of power stations. Specifically, a 110kV station typically has only one 220kV upstream connection but multiple 10kV downstream connections. In contrast, DE performs better than CI in road networks, specifically primary and tertiary roads, due to the degree homogeneity, where road nodes typically have one incoming and one outgoing connection. Comparing the heuristic methods in the coupled network, CI still outperforms DE. We have established a cascade rule where a power station failure leads to the shutdown of traffic lights supported by that station, resulting in congestion and chaos. This demonstrates that power stations have a greater influence in this coupled network.
    Regarding machine-assisted analysis, the method works better than heuristic approaches in road networks because it requires fewer enabled nodes to cover a majority of nodes using a set of samples. However, in the electricity network and coupled network, sampling from such a large number of nodes struggles to cover even half of the total nodes. This poses a challenge due to the significant disparity between the volume of testing data and the available training data, resulting in unsatisfactory performance.
    \item \textbf{Our method achieves the best performance.} 
    In comparison to heuristic methods and other learned machine approaches, our method consistently identifies critical nodes to minimize the current state and maximize the defined reward. By damaging fewer than ten nodes out of a total of 10,887 in the entire graph, our method reduces the electricity power to approximately 50\% and even lower in the case of road networks. The power decrease achieved by our method is 50\% lower than that of CI, the connectivity obtained is approximately 75\% lower in primary roads and about 43\% lower in tertiary roads compared to the best-performing alternative methods. Additionally, the accumulated reward in our process is approximately 37\% higher than that of the CI method.
    \item \textbf{Our method gets the result fast.} 
    Our method efficiently simulates the entire environment and rapidly identifies vulnerable nodes, yielding a stable and effective solution within approximately 500 epochs.
    \item \textbf{Cascade failure exists.} 
    Cascade failure does exist between the topology of interconnectedness between infrastructures. Without prior knowledge, our method could understand and characterize such a cascading rule driven by the data itself and model the complicated effect of evolution mechanisms of urban infrastructures.
\end{itemize}

\subsection{Study on Transferring Ability (RQ2)} \label{tranfer}
As mentioned before, one of the primary purposes is to characterize the vulnerability of the urban system to descend the influence of damaged functional units in cities. However, inaccuracy is always a property of real-world data and the infrastructures will also be continuously built and renovated. Due to the time and computation costs to re-train a new agent finding vulnerable nodes, we try to get the node embedding that responds to the real-world fast and accurately, and leverage the original parameters of value networks to find nodes and protect the current system, to study the transferring ability of our framework \cite{tang2020transferring}. The detailed setup of the transferability experiments is provided in Appendix.
The results have been shown in Figure \ref{fig::transfer}, figure \textbf{(a)-(c)} show the experimental results on the electricity network, primary road, and coupled network, respectively. CI can still find fragile nodes in the electricity and road network. Compared with baselines methods on the new-generated topologies, our transferring system still works better than heuristic methods including DE and CI, obtaining about 46\% lower power decrease and 75\% lower connectivity than the best baseline correspondingly and reaching a plateau after a small set of nodes damaging. In the coupled network, the total reward obtained by our method is about 20 times the reward of DE and 3.4 times the reward of CI. 
 In conclusion, our method has very strong transferring ability and robustness enabling us to avoid huge costs of time and computation. In addition, the improvement is meaningful and evident as urban infrastructures change rapidly in the real world nowadays.

\subsection{Ablation Study (RQ3)}
In reinforcement learning, the agent's objective is to maximize the expectation of the total reward by exploring the environment and taking appropriate actions. As reinforcement learning algorithms become more powerful, agents can handle increasingly complex environments. However, when dealing with heterogeneous graphs consisting of different types of nodes and edges, the exploration-exploitation process for the agent can become challenging. Our hypothesis suggests that providing the agent with a confusing or randomly generated environment, specifically random-generated node embeddings, may result in a longer training process to find a stable solution, and the obtained results may be unsatisfactory.

Using the same datasets and metrics as described in Section \ref{RQ1}, we compare the training duration and final results between our RL agent trained with pre-trained node embeddings generated by GNN and those generated randomly. The comparison is presented in Figure  \ref{fig::abla}, following the same layout as Figure \ref{fig::main}. As indicated in the dataset introduction, the performance of the GNN + RL approach outperforms the learning process in a random environment. Notably, the performance gap between the two approaches is more pronounced in the case of the coupled network compared to individual networks. Additionally, the ablation study on the two road networks shows a subtle difference since road structures are simpler and more limited, allowing for similar performance when the training process is extensive. However, in the interdependent network with multiple types of nodes and edges, learning valuable information from random representations becomes challenging. This results in poor performance during testing, where the total reward plateaus and the metrics of the two interacting single networks show a downward trend.

In conclusion, we have confirmed the validity of our assumption. The inclusion of the GNN component in our framework is essential and significantly enhances the performance of the agent when dealing with graph environments, particularly heterogeneous graphs.
 
\subsection{Case Study (RQ4)}
\begin{figure*}[htbp]
  \centering
  \subfloat[Pre-stage of Electricity Network]{\includegraphics[width=6.cm]{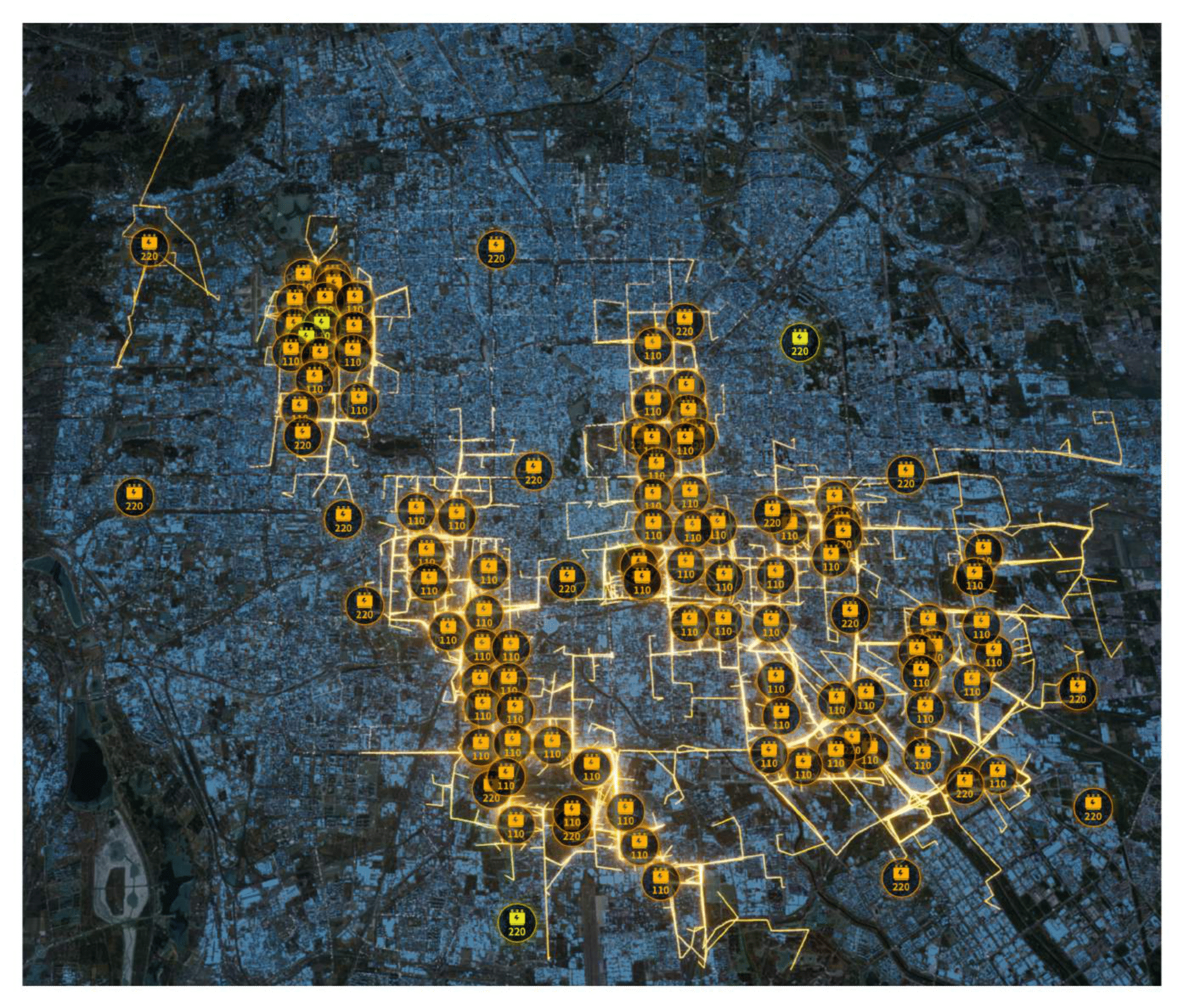}}
  \subfloat[Mid-stage of Electricity Network]{\includegraphics[width=6.cm]{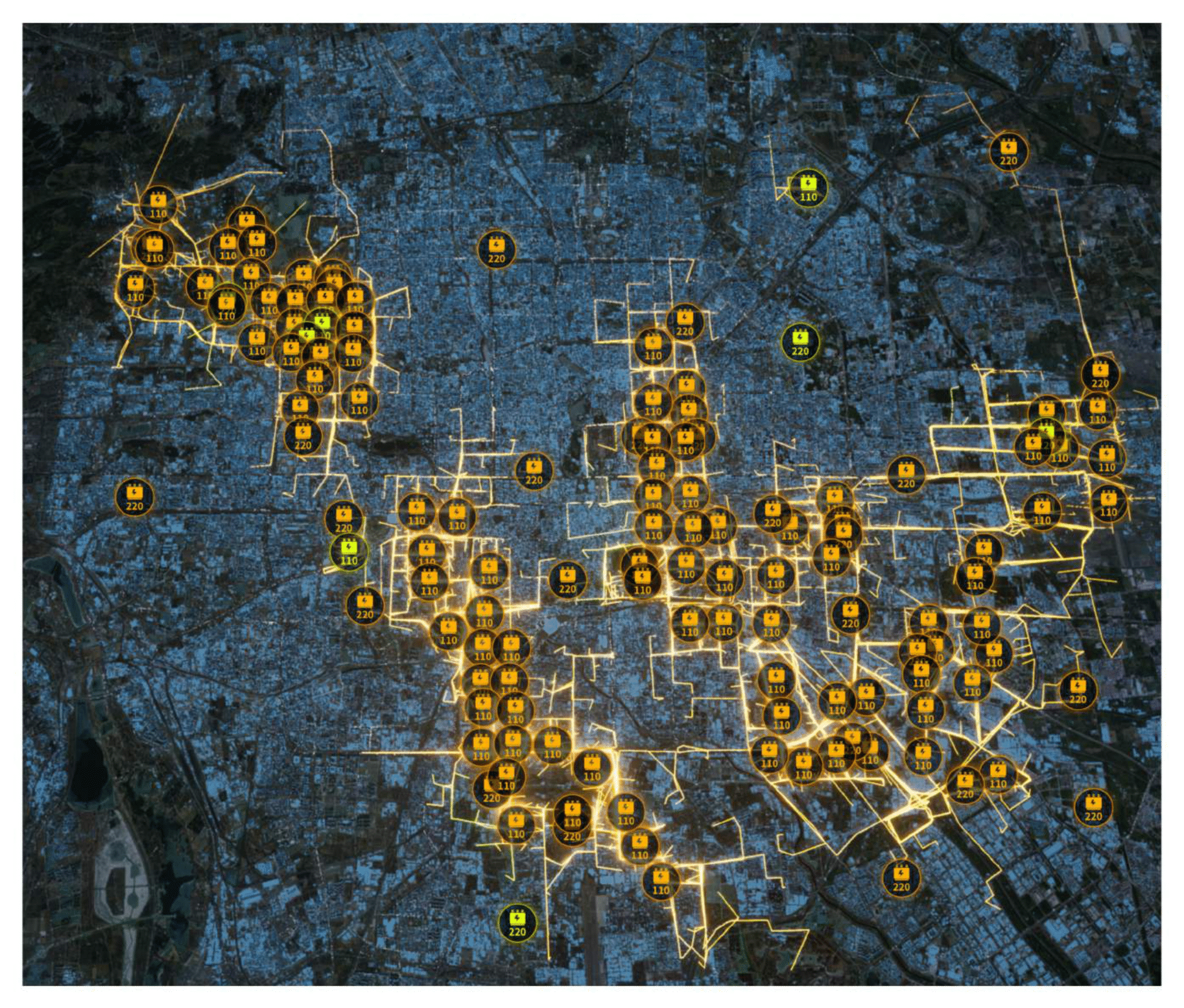}}
  \subfloat[Post-stage of Electricity Network]{\includegraphics[width=6.cm]{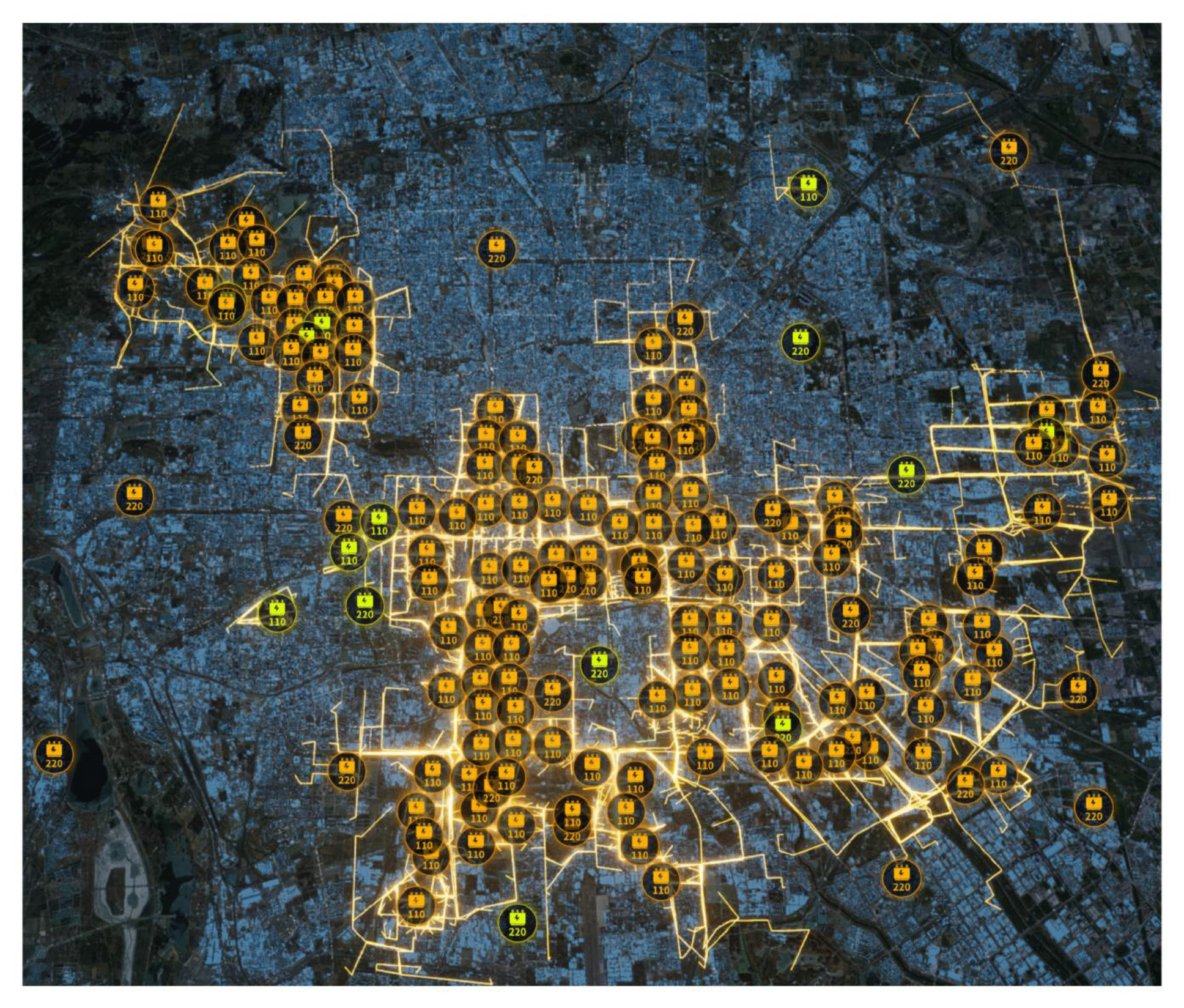}}
  
  \subfloat[Pre-stage of Road Network]{\includegraphics[width=6.cm]{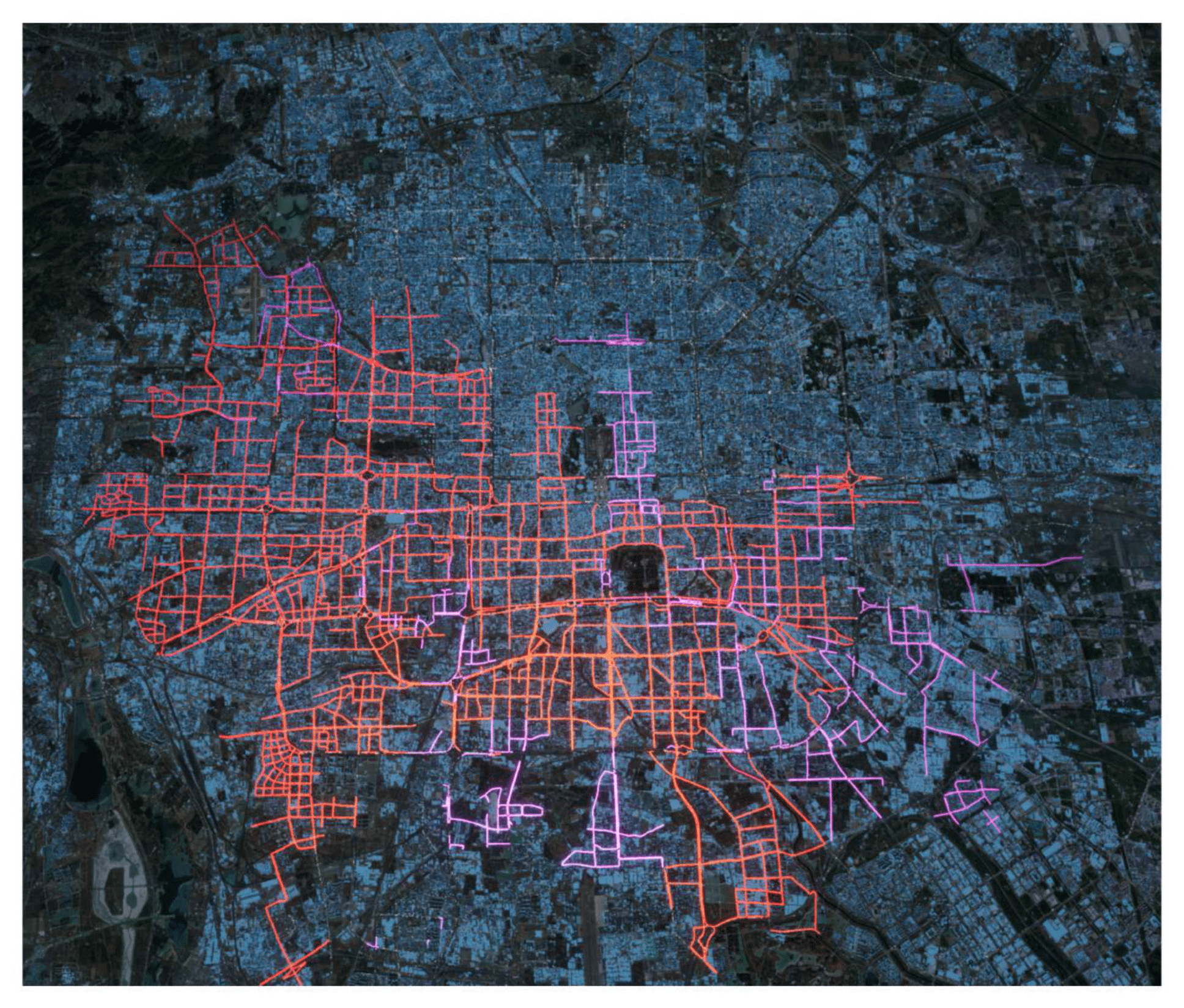}}
  \subfloat[Mid-stage of Road Network]{\includegraphics[width=6.cm]{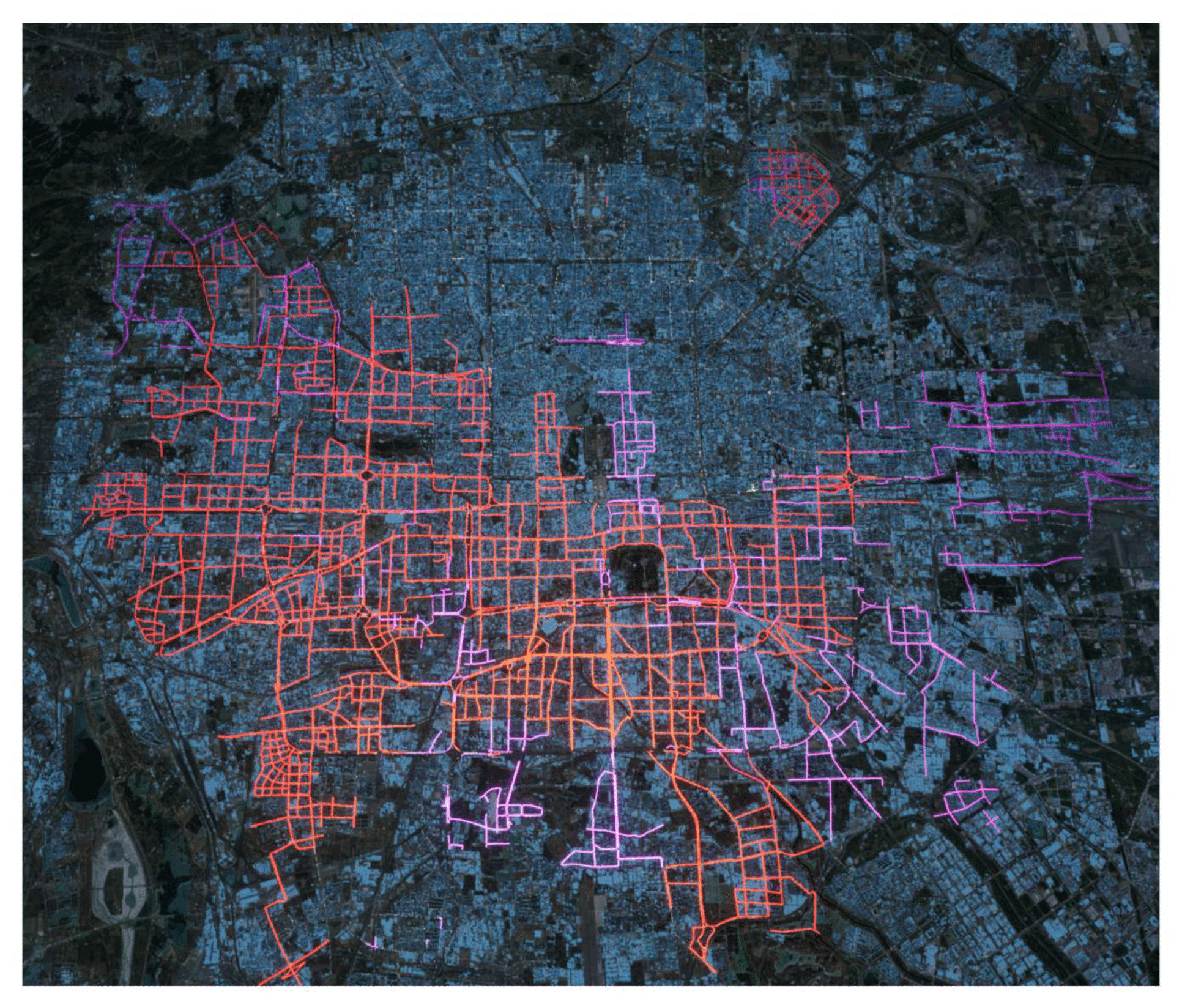}}
  \subfloat[Post-stage of Road Network]{\includegraphics[width=6.cm]{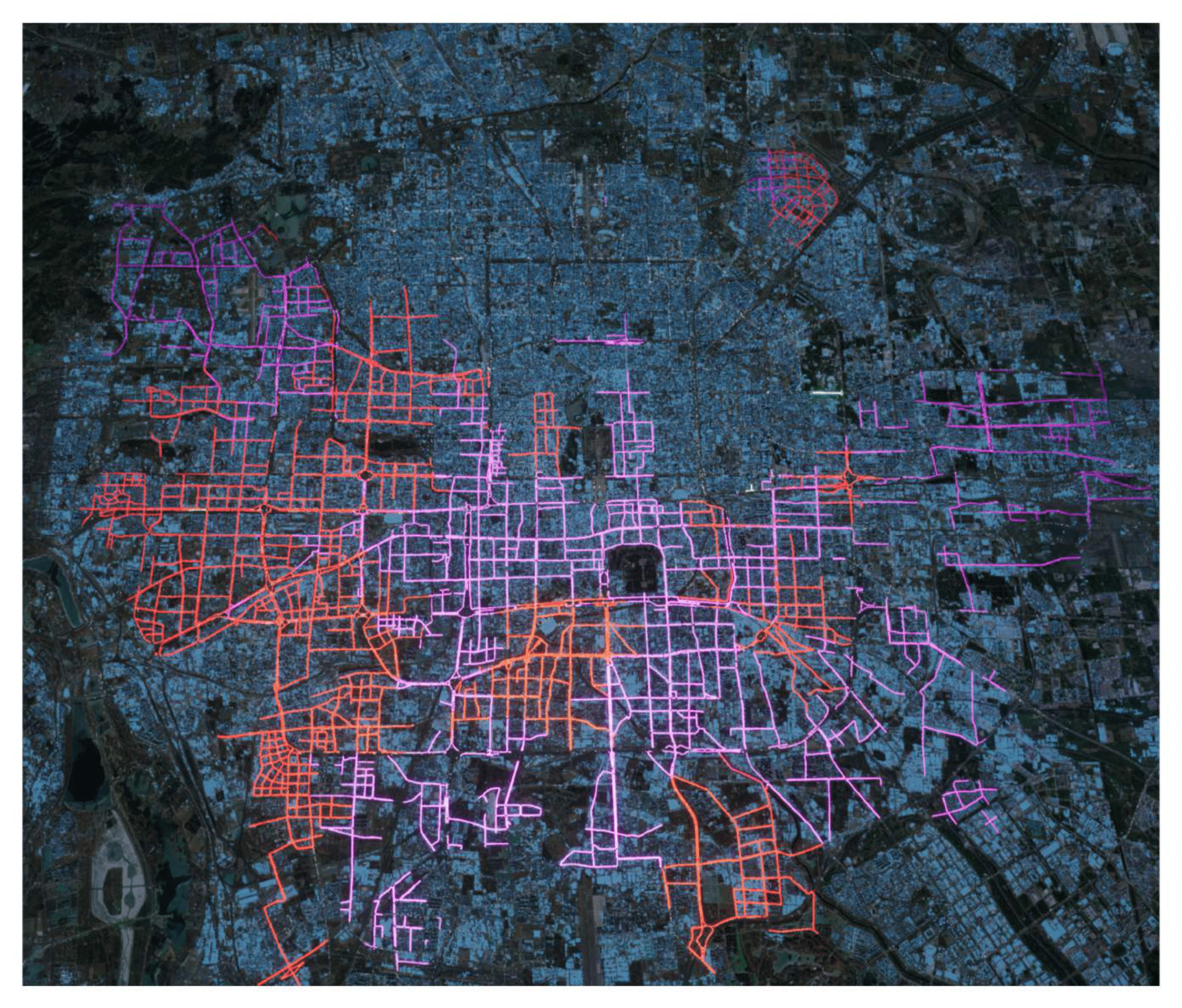}}

  \setlength{\abovecaptionskip}{0.2cm}
        \setlength{\belowcaptionskip}{0.2cm}
  \caption{Process of finding vulnerable nodes in interdependent network} 
  \label{fig::demo}
\end{figure*}

Figure \ref{fig::demo} visualizes the effects of a series of node damages, displaying three sets of images representing the pre-stage, middle stage, and post-stage of identifying vulnerable nodes in the interdependent network \cite{zhang2022mirage}. In the case of the electricity network \textbf{(a)-(c)}, direct destruction is denoted by green markers, while yellow markers represent indirect destruction. For the road network \textbf{(d)-(f)}, light green indicates direct road collapse, magenta represents roads affected by ruined or non-functioning power stations, and red denotes roads separated from the largest connected component after the series of damages.

The impact of node selection in the electricity network is evidently more pronounced in terms of both the extent of influence and power decrease compared to the coupled network. For instance, damaging two nodes in the single electricity network reduces the power to approximately $6.5e9$, whereas in the coupled network, the power remains above $7e9$ even after the same number of nodes are damaged. However, when considering the connectivity in the road network, the topological changes undergo a significant shift as nodes in the electricity network are damaged. As the damage continues, the coverage area of normally operating power stations and the largest connected component become increasingly limited. This demonstrates the existence of cascade failure and validates our framework's ability to accurately characterize the vulnerability of urban infrastructures. Moreover, the number of directly collapsed roads is considerably lower compared to those affected by the electricity network and subsequently isolated. This finding aligns with our model's performance, which shows that the electricity network has a greater influence.

\section{Related Work}\label{sec::relatedwork}

\subsection{Data science in urban infrastructure}
Urban infrastructure, including road networks, communication networks, electricity networks, etc., is one of the most important elements in the city, and there are many applications for data-driven urban computing research.
The early efforts mainly focus on the modeling or prediction of urban infrastructure.
For road networks, the existing works pay a lot of attention to the problem of road traffic forcasting~\cite{deng2016latent}, road traffic 
analaysis~\cite{liu2011discovering}, travel time prediction~\cite{wang2014travel}, or using the road network to predict other elements in the city such as air quality~\cite{yuan2012discovering}.
That is, the existing works are still concentrated on one specific kind of infrastructure network, ignoring the joint modeling of the interdependence relations among different networks, which we aim to address in this work.

Despite the early efforts, the resilience and vulnerability of urban infrastructure are less explored in the literature. In this work, we approach the problem of finding the critical node in the interdependent infrastructure network, with extensive applications in the real world.

\subsection{Data-driven optimization on graph}
Graphs widely exist in plenty of domains, such as social networks, user-item graphs in recommender systems, road networks, etc.
The ubiquity of graphs makes the optimization task of graphs an important problem in many real-world applications.
The typical optimization problems include the traveling salesman problem (TSP)~\cite{junger1995traveling}, maximum independent set (MIS)~\cite{tarjan1977finding}, maximum cut (MaxCut)~\cite{festa2002randomized}, etc.
However, despite their importance and wide applications, these optimization problems are always NP-hard, leading to high-cost time complexity that is exponentially related to the graph size.
The traditional methods to address it are based on heuristics guided by real-world experience from human experts.
However, although heuristic methods may not perform badly, they still suffer from limitations, which can be resolved by data-driven machine learning models~\cite{bengio2021machine}.
First, the heuristic methods highly rely on problem-specific designs, which will be challenging for new scenarios.
Second, the data-driven methods can extract useful features or prediction signals that may be implicit and hard to explain, making them less likely to be proposed by human experts.
Khalil~\textit{et al.}~\cite{khalil2017learning} proposed to learn graph states with graph neural networks and greedily generate actions until the criteria are met, supporting the data-driven solutions for Minimum Vertex Cover, Maximum Cut, and Traveling Salesman problems.
Li~\textit{et al.}~\cite{li2018combinatorial} propose to use a graph convolutional network to predict whether a given vertex belongs to the optimal set or not, converting the original problem to a supervised-learning task. The authors further combine the tree search method with the neural network module to ensure the exploration ability of the approach. 

Different from the existing works which only consider small graphs, in this work, we approach a real-world problem in the urban infrastructure with far larger graph and heterogeneous relations, along with the complex environment.

\section{Conclusion}\label{sec::conclusion}
Addressing the vulnerability of urban infrastructure interdependent network, we propose a data-driven framework using graph neural network and reinforcement learning to detect vulnerable nodes in urban infrastructure networks. Requiring no prior expert knowledge but just the connected relationships between different types of nodes in the real world, it trains on the complicated network quite efficiently. Thanks to the transferring ability and robustness of our framework, we can ignore some slight perturbations in our data and trust our model's performance in terms of effectiveness. Besides, our framework also illustrates the necessity of an explicit representation for the complicated network.
Our framework captures the risk of cascade failure and discovers vulnerable nodes in real-world systems, which turns out to be dramatically useful in protecting fragile nodes in urban infrastructure and designing more robust topologies for cities.

\begin{acks}
    This work is supported by the National Key Research and Development Program of China (2022YFF0606904), the National Natural Science Foundation of China (U22B2057, 62171260, and 62272262), the Beijing National Research Center for Information Science and Technology, and the China Postdoctoral Science Foundation fellowship (2021TQ0027 and 2022M710006).
\end{acks}

\clearpage
\bibliographystyle{ACM-Reference-Format}
\balance
\bibliography{bibliography}


\begin{thebibliography}{40}


\ifx \showCODEN    \undefined \def \showCODEN     #1{\unskip}     \fi
\ifx \showDOI      \undefined \def \showDOI       #1{#1}\fi
\ifx \showISBNx    \undefined \def \showISBNx     #1{\unskip}     \fi
\ifx \showISBNxiii \undefined \def \showISBNxiii  #1{\unskip}     \fi
\ifx \showISSN     \undefined \def \showISSN      #1{\unskip}     \fi
\ifx \showLCCN     \undefined \def \showLCCN      #1{\unskip}     \fi
\ifx \shownote     \undefined \def \shownote      #1{#1}          \fi
\ifx \showarticletitle \undefined \def \showarticletitle #1{#1}   \fi
\ifx \showURL      \undefined \def \showURL       {\relax}        \fi
\providecommand\bibfield[2]{#2}
\providecommand\bibinfo[2]{#2}
\providecommand\natexlab[1]{#1}
\providecommand\showeprint[2][]{arXiv:#2}

\bibitem[\protect\citeauthoryear{Abdulla and Birgisson}{Abdulla and
  Birgisson}{2020}]%
        {abdulla2020predicting}
\bibfield{author}{\bibinfo{person}{Bahrulla Abdulla} {and}
  \bibinfo{person}{Bjorn Birgisson}.} \bibinfo{year}{2020}\natexlab{}.
\newblock \showarticletitle{Predicting road network vulnerability to fluvial
  flooding using machine learning classifiers: Case study of Houston during
  Hurricane Harvey}. In \bibinfo{booktitle}{{\em Construction Research Congress
  2020: Computer Applications}}. \bibinfo{pages}{38--47}.
\newblock


\bibitem[\protect\citeauthoryear{Bengio, Lodi, and Prouvost}{Bengio
  et~al\mbox{.}}{2021}]%
        {bengio2021machine}
\bibfield{author}{\bibinfo{person}{Yoshua Bengio}, \bibinfo{person}{Andrea
  Lodi}, {and} \bibinfo{person}{Antoine Prouvost}.}
  \bibinfo{year}{2021}\natexlab{}.
\newblock \showarticletitle{Machine learning for combinatorial optimization: a
  methodological tour d’horizon}.
\newblock \bibinfo{journal}{{\em European Journal of Operational Research\/}}
  \bibinfo{volume}{290}, \bibinfo{number}{2} (\bibinfo{year}{2021}),
  \bibinfo{pages}{405--421}.
\newblock


\bibitem[\protect\citeauthoryear{Brummitt, D’Souza, and Leicht}{Brummitt
  et~al\mbox{.}}{2012}]%
        {brummitt2012suppressing}
\bibfield{author}{\bibinfo{person}{Charles~D Brummitt},
  \bibinfo{person}{Raissa~M D’Souza}, {and} \bibinfo{person}{Elizabeth~A
  Leicht}.} \bibinfo{year}{2012}\natexlab{}.
\newblock \showarticletitle{Suppressing cascades of load in interdependent
  networks}.
\newblock \bibinfo{journal}{{\em Proceedings of the national academy of
  sciences\/}} \bibinfo{volume}{109}, \bibinfo{number}{12}
  (\bibinfo{year}{2012}), \bibinfo{pages}{E680--E689}.
\newblock


\bibitem[\protect\citeauthoryear{Buldyrev, Parshani, Paul, Stanley, and
  Havlin}{Buldyrev et~al\mbox{.}}{2010}]%
        {buldyrev2010catastrophic}
\bibfield{author}{\bibinfo{person}{Sergey~V Buldyrev}, \bibinfo{person}{Roni
  Parshani}, \bibinfo{person}{Gerald Paul}, \bibinfo{person}{H~Eugene Stanley},
  {and} \bibinfo{person}{Shlomo Havlin}.} \bibinfo{year}{2010}\natexlab{}.
\newblock \showarticletitle{Catastrophic cascade of failures in interdependent
  networks}.
\newblock \bibinfo{journal}{{\em Nature\/}} \bibinfo{volume}{464},
  \bibinfo{number}{7291} (\bibinfo{year}{2010}), \bibinfo{pages}{1025--1028}.
\newblock


\bibitem[\protect\citeauthoryear{Cohen, Erez, Ben-Avraham, and Havlin}{Cohen
  et~al\mbox{.}}{2000}]%
        {cohen2000resilience}
\bibfield{author}{\bibinfo{person}{Reuven Cohen}, \bibinfo{person}{Keren Erez},
  \bibinfo{person}{Daniel Ben-Avraham}, {and} \bibinfo{person}{Shlomo Havlin}.}
  \bibinfo{year}{2000}\natexlab{}.
\newblock \showarticletitle{Resilience of the internet to random breakdowns}.
\newblock \bibinfo{journal}{{\em Physical review letters\/}}
  \bibinfo{volume}{85}, \bibinfo{number}{21} (\bibinfo{year}{2000}),
  \bibinfo{pages}{4626}.
\newblock


\bibitem[\protect\citeauthoryear{Collier and Venables}{Collier and
  Venables}{2016}]%
        {collier2016urban}
\bibfield{author}{\bibinfo{person}{Paul Collier} {and}
  \bibinfo{person}{Anthony~J Venables}.} \bibinfo{year}{2016}\natexlab{}.
\newblock \showarticletitle{Urban infrastructure for development}.
\newblock \bibinfo{journal}{{\em Oxford Review of Economic Policy\/}}
  \bibinfo{volume}{32}, \bibinfo{number}{3} (\bibinfo{year}{2016}),
  \bibinfo{pages}{391--409}.
\newblock


\bibitem[\protect\citeauthoryear{Darong, Libing, and Ling}{Darong
  et~al\mbox{.}}{2015}]%
        {darong2015vulnerability}
\bibfield{author}{\bibinfo{person}{Huang Darong}, \bibinfo{person}{Shen
  Libing}, {and} \bibinfo{person}{Zhao Ling}.} \bibinfo{year}{2015}\natexlab{}.
\newblock \showarticletitle{Vulnerability analysis of urban road network based
  on complex network theory}.
\newblock \bibinfo{journal}{{\em Journal of Chongqing Jiaotong University
  (Natural Science)\/}} \bibinfo{volume}{34}, \bibinfo{number}{1}
  (\bibinfo{year}{2015}), \bibinfo{pages}{110}.
\newblock


\bibitem[\protect\citeauthoryear{Deng, Shahabi, Demiryurek, Zhu, Yu, and
  Liu}{Deng et~al\mbox{.}}{2016}]%
        {deng2016latent}
\bibfield{author}{\bibinfo{person}{Dingxiong Deng}, \bibinfo{person}{Cyrus
  Shahabi}, \bibinfo{person}{Ugur Demiryurek}, \bibinfo{person}{Linhong Zhu},
  \bibinfo{person}{Rose Yu}, {and} \bibinfo{person}{Yan Liu}.}
  \bibinfo{year}{2016}\natexlab{}.
\newblock \showarticletitle{Latent space model for road networks to predict
  time-varying traffic}. In \bibinfo{booktitle}{{\em Proceedings of the 22nd
  ACM SIGKDD international conference on knowledge discovery and data mining}}.
  \bibinfo{pages}{1525--1534}.
\newblock


\bibitem[\protect\citeauthoryear{Dunn, Wilkinson, and Ford}{Dunn
  et~al\mbox{.}}{2016}]%
        {dunn2016spatial}
\bibfield{author}{\bibinfo{person}{Sarah Dunn}, \bibinfo{person}{Sean
  Wilkinson}, {and} \bibinfo{person}{Alistair Ford}.}
  \bibinfo{year}{2016}\natexlab{}.
\newblock \showarticletitle{Spatial structure and evolution of infrastructure
  networks}.
\newblock \bibinfo{journal}{{\em Sustainable cities and society\/}}
  \bibinfo{volume}{27} (\bibinfo{year}{2016}), \bibinfo{pages}{23--31}.
\newblock


\bibitem[\protect\citeauthoryear{Eusgeld, Kr{\"o}ger, Sansavini, Schl{\"a}pfer,
  and Zio}{Eusgeld et~al\mbox{.}}{2009}]%
        {eusgeld2009role}
\bibfield{author}{\bibinfo{person}{Irene Eusgeld}, \bibinfo{person}{Wolfgang
  Kr{\"o}ger}, \bibinfo{person}{Giovanni Sansavini}, \bibinfo{person}{Markus
  Schl{\"a}pfer}, {and} \bibinfo{person}{Enrico Zio}.}
  \bibinfo{year}{2009}\natexlab{}.
\newblock \showarticletitle{The role of network theory and object-oriented
  modeling within a framework for the vulnerability analysis of critical
  infrastructures}.
\newblock \bibinfo{journal}{{\em Reliability Engineering \& System Safety\/}}
  \bibinfo{volume}{94}, \bibinfo{number}{5} (\bibinfo{year}{2009}),
  \bibinfo{pages}{954--963}.
\newblock


\bibitem[\protect\citeauthoryear{Fan, Zeng, Sun, and Liu}{Fan
  et~al\mbox{.}}{2020}]%
        {fan2020finding}
\bibfield{author}{\bibinfo{person}{Changjun Fan}, \bibinfo{person}{Li Zeng},
  \bibinfo{person}{Yizhou Sun}, {and} \bibinfo{person}{Yang-Yu Liu}.}
  \bibinfo{year}{2020}\natexlab{}.
\newblock \showarticletitle{Finding key players in complex networks through
  deep reinforcement learning}.
\newblock \bibinfo{journal}{{\em Nature machine intelligence\/}}
  \bibinfo{volume}{2}, \bibinfo{number}{6} (\bibinfo{year}{2020}),
  \bibinfo{pages}{317--324}.
\newblock


\bibitem[\protect\citeauthoryear{Festa, Pardalos, Resende, and Ribeiro}{Festa
  et~al\mbox{.}}{2002}]%
        {festa2002randomized}
\bibfield{author}{\bibinfo{person}{Paola Festa}, \bibinfo{person}{Panos~M
  Pardalos}, \bibinfo{person}{Mauricio~GC Resende}, {and}
  \bibinfo{person}{Celso~C Ribeiro}.} \bibinfo{year}{2002}\natexlab{}.
\newblock \showarticletitle{Randomized heuristics for the MAX-CUT problem}.
\newblock \bibinfo{journal}{{\em Optimization methods and software\/}}
  \bibinfo{volume}{17}, \bibinfo{number}{6} (\bibinfo{year}{2002}),
  \bibinfo{pages}{1033--1058}.
\newblock


\bibitem[\protect\citeauthoryear{Gao, Zheng, Li, Li, Qin, Piao, Quan, Chang,
  Jin, He, et~al\mbox{.}}{Gao et~al\mbox{.}}{2023}]%
        {gao2023survey}
\bibfield{author}{\bibinfo{person}{Chen Gao}, \bibinfo{person}{Yu Zheng},
  \bibinfo{person}{Nian Li}, \bibinfo{person}{Yinfeng Li},
  \bibinfo{person}{Yingrong Qin}, \bibinfo{person}{Jinghua Piao},
  \bibinfo{person}{Yuhan Quan}, \bibinfo{person}{Jianxin Chang},
  \bibinfo{person}{Depeng Jin}, \bibinfo{person}{Xiangnan He}, {et~al\mbox{.}}}
  \bibinfo{year}{2023}\natexlab{}.
\newblock \showarticletitle{A survey of graph neural networks for recommender
  systems: challenges, methods, and directions}.
\newblock \bibinfo{journal}{{\em ACM Transactions on Recommender Systems\/}}
  \bibinfo{volume}{1}, \bibinfo{number}{1} (\bibinfo{year}{2023}),
  \bibinfo{pages}{1--51}.
\newblock


\bibitem[\protect\citeauthoryear{Gao, Barzel, and Barab{\'a}si}{Gao
  et~al\mbox{.}}{2016}]%
        {gao2016universal}
\bibfield{author}{\bibinfo{person}{Jianxi Gao}, \bibinfo{person}{Baruch
  Barzel}, {and} \bibinfo{person}{Albert-L{\'a}szl{\'o} Barab{\'a}si}.}
  \bibinfo{year}{2016}\natexlab{}.
\newblock \showarticletitle{Universal resilience patterns in complex networks}.
\newblock \bibinfo{journal}{{\em Nature\/}} \bibinfo{volume}{530},
  \bibinfo{number}{7590} (\bibinfo{year}{2016}), \bibinfo{pages}{307--312}.
\newblock


\bibitem[\protect\citeauthoryear{Grassia, De~Domenico, and Mangioni}{Grassia
  et~al\mbox{.}}{2021}]%
        {grassia2021machine}
\bibfield{author}{\bibinfo{person}{Marco Grassia}, \bibinfo{person}{Manlio
  De~Domenico}, {and} \bibinfo{person}{Giuseppe Mangioni}.}
  \bibinfo{year}{2021}\natexlab{}.
\newblock \showarticletitle{Machine learning dismantling and early-warning
  signals of disintegration in complex systems}.
\newblock \bibinfo{journal}{{\em Nature Communications\/}}
  \bibinfo{volume}{12}, \bibinfo{number}{1} (\bibinfo{year}{2021}),
  \bibinfo{pages}{5190}.
\newblock


\bibitem[\protect\citeauthoryear{Grover and Leskovec}{Grover and
  Leskovec}{2016}]%
        {grover2016node2vec}
\bibfield{author}{\bibinfo{person}{Aditya Grover} {and} \bibinfo{person}{Jure
  Leskovec}.} \bibinfo{year}{2016}\natexlab{}.
\newblock \showarticletitle{node2vec: Scalable feature learning for networks}.
  In \bibinfo{booktitle}{{\em Proceedings of the 22nd ACM SIGKDD international
  conference on Knowledge discovery and data mining}}.
  \bibinfo{pages}{855--864}.
\newblock


\bibitem[\protect\citeauthoryear{Hessel, Modayil, Van~Hasselt, Schaul,
  Ostrovski, Dabney, Horgan, Piot, Azar, and Silver}{Hessel
  et~al\mbox{.}}{2018}]%
        {hessel2018rainbow}
\bibfield{author}{\bibinfo{person}{Matteo Hessel}, \bibinfo{person}{Joseph
  Modayil}, \bibinfo{person}{Hado Van~Hasselt}, \bibinfo{person}{Tom Schaul},
  \bibinfo{person}{Georg Ostrovski}, \bibinfo{person}{Will Dabney},
  \bibinfo{person}{Dan Horgan}, \bibinfo{person}{Bilal Piot},
  \bibinfo{person}{Mohammad Azar}, {and} \bibinfo{person}{David Silver}.}
  \bibinfo{year}{2018}\natexlab{}.
\newblock \showarticletitle{Rainbow: Combining improvements in deep
  reinforcement learning}. In \bibinfo{booktitle}{{\em Proceedings of the AAAI
  conference on artificial intelligence}}, Vol.~\bibinfo{volume}{32}.
\newblock


\bibitem[\protect\citeauthoryear{Hines, Cotilla-Sanchez, and Blumsack}{Hines
  et~al\mbox{.}}{2010}]%
        {hines2010topological}
\bibfield{author}{\bibinfo{person}{Paul Hines}, \bibinfo{person}{Eduardo
  Cotilla-Sanchez}, {and} \bibinfo{person}{Seth Blumsack}.}
  \bibinfo{year}{2010}\natexlab{}.
\newblock \showarticletitle{Do topological models provide good information
  about electricity infrastructure vulnerability?}
\newblock \bibinfo{journal}{{\em Chaos: An Interdisciplinary Journal of
  Nonlinear Science\/}} \bibinfo{volume}{20}, \bibinfo{number}{3}
  (\bibinfo{year}{2010}), \bibinfo{pages}{033122}.
\newblock


\bibitem[\protect\citeauthoryear{Iyer, Killingback, Sundaram, and Wang}{Iyer
  et~al\mbox{.}}{2013}]%
        {iyer2013attack}
\bibfield{author}{\bibinfo{person}{Swami Iyer}, \bibinfo{person}{Timothy
  Killingback}, \bibinfo{person}{Bala Sundaram}, {and} \bibinfo{person}{Zhen
  Wang}.} \bibinfo{year}{2013}\natexlab{}.
\newblock \showarticletitle{Attack robustness and centrality of complex
  networks}.
\newblock \bibinfo{journal}{{\em PloS one\/}} \bibinfo{volume}{8},
  \bibinfo{number}{4} (\bibinfo{year}{2013}), \bibinfo{pages}{e59613}.
\newblock


\bibitem[\protect\citeauthoryear{J{\"u}nger, Reinelt, and Rinaldi}{J{\"u}nger
  et~al\mbox{.}}{1995}]%
        {junger1995traveling}
\bibfield{author}{\bibinfo{person}{Michael J{\"u}nger},
  \bibinfo{person}{Gerhard Reinelt}, {and} \bibinfo{person}{Giovanni Rinaldi}.}
  \bibinfo{year}{1995}\natexlab{}.
\newblock \showarticletitle{The traveling salesman problem}.
\newblock \bibinfo{journal}{{\em Handbooks in operations research and
  management science\/}}  \bibinfo{volume}{7} (\bibinfo{year}{1995}),
  \bibinfo{pages}{225--330}.
\newblock


\bibitem[\protect\citeauthoryear{Khalil, Dai, Zhang, Dilkina, and Song}{Khalil
  et~al\mbox{.}}{2017}]%
        {khalil2017learning}
\bibfield{author}{\bibinfo{person}{Elias Khalil}, \bibinfo{person}{Hanjun Dai},
  \bibinfo{person}{Yuyu Zhang}, \bibinfo{person}{Bistra Dilkina}, {and}
  \bibinfo{person}{Le Song}.} \bibinfo{year}{2017}\natexlab{}.
\newblock \showarticletitle{Learning combinatorial optimization algorithms over
  graphs}.
\newblock \bibinfo{journal}{{\em Advances in neural information processing
  systems\/}}  \bibinfo{volume}{30} (\bibinfo{year}{2017}).
\newblock


\bibitem[\protect\citeauthoryear{Kipf and Welling}{Kipf and Welling}{2016}]%
        {kipf2016semi}
\bibfield{author}{\bibinfo{person}{Thomas~N Kipf} {and} \bibinfo{person}{Max
  Welling}.} \bibinfo{year}{2016}\natexlab{}.
\newblock \showarticletitle{Semi-supervised classification with graph
  convolutional networks}.
\newblock \bibinfo{journal}{{\em arXiv preprint arXiv:1609.02907\/}}
  (\bibinfo{year}{2016}).
\newblock


\bibitem[\protect\citeauthoryear{Li, Shang, and Yang}{Li et~al\mbox{.}}{2017}]%
        {li2017clustering}
\bibfield{author}{\bibinfo{person}{Yusheng Li}, \bibinfo{person}{Yilun Shang},
  {and} \bibinfo{person}{Yiting Yang}.} \bibinfo{year}{2017}\natexlab{}.
\newblock \showarticletitle{Clustering coefficients of large networks}.
\newblock \bibinfo{journal}{{\em Information Sciences\/}}
  \bibinfo{volume}{382} (\bibinfo{year}{2017}), \bibinfo{pages}{350--358}.
\newblock


\bibitem[\protect\citeauthoryear{Li, Chen, and Koltun}{Li
  et~al\mbox{.}}{2018}]%
        {li2018combinatorial}
\bibfield{author}{\bibinfo{person}{Zhuwen Li}, \bibinfo{person}{Qifeng Chen},
  {and} \bibinfo{person}{Vladlen Koltun}.} \bibinfo{year}{2018}\natexlab{}.
\newblock \showarticletitle{Combinatorial optimization with graph convolutional
  networks and guided tree search}.
\newblock \bibinfo{journal}{{\em Advances in neural information processing
  systems\/}}  \bibinfo{volume}{31} (\bibinfo{year}{2018}).
\newblock


\bibitem[\protect\citeauthoryear{Liu, Zheng, Chawla, Yuan, and Xing}{Liu
  et~al\mbox{.}}{2011}]%
        {liu2011discovering}
\bibfield{author}{\bibinfo{person}{Wei Liu}, \bibinfo{person}{Yu Zheng},
  \bibinfo{person}{Sanjay Chawla}, \bibinfo{person}{Jing Yuan}, {and}
  \bibinfo{person}{Xie Xing}.} \bibinfo{year}{2011}\natexlab{}.
\newblock \showarticletitle{Discovering spatio-temporal causal interactions in
  traffic data streams}. In \bibinfo{booktitle}{{\em Proceedings of the 17th
  ACM SIGKDD international conference on Knowledge discovery and data mining}}.
  \bibinfo{pages}{1010--1018}.
\newblock


\bibitem[\protect\citeauthoryear{Mnih, Kavukcuoglu, Silver, Rusu, Veness,
  Bellemare, Graves, Riedmiller, Fidjeland, Ostrovski, et~al\mbox{.}}{Mnih
  et~al\mbox{.}}{2015}]%
        {mnih2015human}
\bibfield{author}{\bibinfo{person}{Volodymyr Mnih}, \bibinfo{person}{Koray
  Kavukcuoglu}, \bibinfo{person}{David Silver}, \bibinfo{person}{Andrei~A
  Rusu}, \bibinfo{person}{Joel Veness}, \bibinfo{person}{Marc~G Bellemare},
  \bibinfo{person}{Alex Graves}, \bibinfo{person}{Martin Riedmiller},
  \bibinfo{person}{Andreas~K Fidjeland}, \bibinfo{person}{Georg Ostrovski},
  {et~al\mbox{.}}} \bibinfo{year}{2015}\natexlab{}.
\newblock \showarticletitle{Human-level control through deep reinforcement
  learning}.
\newblock \bibinfo{journal}{{\em nature\/}} \bibinfo{volume}{518},
  \bibinfo{number}{7540} (\bibinfo{year}{2015}), \bibinfo{pages}{529--533}.
\newblock


\bibitem[\protect\citeauthoryear{Morone and Makse}{Morone and Makse}{2015}]%
        {morone2015influence}
\bibfield{author}{\bibinfo{person}{Flaviano Morone} {and}
  \bibinfo{person}{Hern{\'a}n~A Makse}.} \bibinfo{year}{2015}\natexlab{}.
\newblock \showarticletitle{Influence maximization in complex networks through
  optimal percolation}.
\newblock \bibinfo{journal}{{\em Nature\/}} \bibinfo{volume}{524},
  \bibinfo{number}{7563} (\bibinfo{year}{2015}), \bibinfo{pages}{65--68}.
\newblock


\bibitem[\protect\citeauthoryear{Ren, Liu, Li, Zhang, Wang, Hao, and Cui}{Ren
  et~al\mbox{.}}{2022}]%
        {ren2022topological}
\bibfield{author}{\bibinfo{person}{Hancheng Ren}, \bibinfo{person}{Shu Liu},
  \bibinfo{person}{Min Li}, \bibinfo{person}{Hongping Zhang},
  \bibinfo{person}{Huiying Wang}, \bibinfo{person}{Xiaoli Hao}, {and}
  \bibinfo{person}{Jie Cui}.} \bibinfo{year}{2022}\natexlab{}.
\newblock \showarticletitle{Topological Analysis and Application of Urban
  Drainage Network}.
\newblock \bibinfo{journal}{{\em Water\/}} \bibinfo{volume}{14},
  \bibinfo{number}{22} (\bibinfo{year}{2022}), \bibinfo{pages}{3732}.
\newblock


\bibitem[\protect\citeauthoryear{Ren, Song, Yang, Baptista, and Grebogi}{Ren
  et~al\mbox{.}}{2016}]%
        {ren2016cascade}
\bibfield{author}{\bibinfo{person}{Hai-Peng Ren}, \bibinfo{person}{Jihong
  Song}, \bibinfo{person}{Rong Yang}, \bibinfo{person}{Murilo~S Baptista},
  {and} \bibinfo{person}{Celso Grebogi}.} \bibinfo{year}{2016}\natexlab{}.
\newblock \showarticletitle{Cascade failure analysis of power grid using new
  load distribution law and node removal rule}.
\newblock \bibinfo{journal}{{\em Physica A: Statistical Mechanics and its
  Applications\/}}  \bibinfo{volume}{442} (\bibinfo{year}{2016}),
  \bibinfo{pages}{239--251}.
\newblock


\bibitem[\protect\citeauthoryear{Rosset, Zhu, and Hastie}{Rosset
  et~al\mbox{.}}{2003}]%
        {rosset2003margin}
\bibfield{author}{\bibinfo{person}{Saharon Rosset}, \bibinfo{person}{Ji Zhu},
  {and} \bibinfo{person}{Trevor Hastie}.} \bibinfo{year}{2003}\natexlab{}.
\newblock \showarticletitle{Margin maximizing loss functions}.
\newblock \bibinfo{journal}{{\em Advances in neural information processing
  systems\/}}  \bibinfo{volume}{16} (\bibinfo{year}{2003}).
\newblock


\bibitem[\protect\citeauthoryear{Sutton and Barto}{Sutton and Barto}{2018}]%
        {sutton2018reinforcement}
\bibfield{author}{\bibinfo{person}{Richard~S Sutton} {and}
  \bibinfo{person}{Andrew~G Barto}.} \bibinfo{year}{2018}\natexlab{}.
\newblock \bibinfo{booktitle}{{\em Reinforcement learning: An introduction}}.
\newblock \bibinfo{publisher}{MIT press}.
\newblock


\bibitem[\protect\citeauthoryear{Szepesvari}{Szepesvari}{2010}]%
        {szepesvari2010algorithms}
\bibfield{author}{\bibinfo{person}{Csaba Szepesvari}.}
  \bibinfo{year}{2010}\natexlab{}.
\newblock \showarticletitle{Algorithms for reinforcement learning: Synthesis
  lectures on artificial intelligence and machine learning}.
\newblock \bibinfo{journal}{{\em Morgan and Claypool\/}}
  (\bibinfo{year}{2010}).
\newblock


\bibitem[\protect\citeauthoryear{Tang, Li, Sun, Yao, Mitra, and Wang}{Tang
  et~al\mbox{.}}{2020}]%
        {tang2020transferring}
\bibfield{author}{\bibinfo{person}{Xianfeng Tang}, \bibinfo{person}{Yandong
  Li}, \bibinfo{person}{Yiwei Sun}, \bibinfo{person}{Huaxiu Yao},
  \bibinfo{person}{Prasenjit Mitra}, {and} \bibinfo{person}{Suhang Wang}.}
  \bibinfo{year}{2020}\natexlab{}.
\newblock \showarticletitle{Transferring robustness for graph neural network
  against poisoning attacks}. In \bibinfo{booktitle}{{\em Proceedings of the
  13th international conference on web search and data mining}}.
  \bibinfo{pages}{600--608}.
\newblock


\bibitem[\protect\citeauthoryear{Tarjan and Trojanowski}{Tarjan and
  Trojanowski}{1977}]%
        {tarjan1977finding}
\bibfield{author}{\bibinfo{person}{Robert~Endre Tarjan} {and}
  \bibinfo{person}{Anthony~E Trojanowski}.} \bibinfo{year}{1977}\natexlab{}.
\newblock \showarticletitle{Finding a maximum independent set}.
\newblock \bibinfo{journal}{{\it SIAM J. Comput.}} \bibinfo{volume}{6},
  \bibinfo{number}{3} (\bibinfo{year}{1977}), \bibinfo{pages}{537--546}.
\newblock


\bibitem[\protect\citeauthoryear{Wang, Zheng, Ye, Gan, Li, Song, Zhou, Ma, Yu,
  Gai, Xiao, He, Karypis, Li, and Zhang}{Wang et~al\mbox{.}}{2019}]%
        {wang2019dgl}
\bibfield{author}{\bibinfo{person}{Minjie Wang}, \bibinfo{person}{Da Zheng},
  \bibinfo{person}{Zihao Ye}, \bibinfo{person}{Quan Gan},
  \bibinfo{person}{Mufei Li}, \bibinfo{person}{Xiang Song},
  \bibinfo{person}{Jinjing Zhou}, \bibinfo{person}{Chao Ma},
  \bibinfo{person}{Lingfan Yu}, \bibinfo{person}{Yu Gai},
  \bibinfo{person}{Tianjun Xiao}, \bibinfo{person}{Tong He},
  \bibinfo{person}{George Karypis}, \bibinfo{person}{Jinyang Li}, {and}
  \bibinfo{person}{Zheng Zhang}.} \bibinfo{year}{2019}\natexlab{}.
\newblock \showarticletitle{Deep Graph Library: A Graph-Centric,
  Highly-Performant Package for Graph Neural Networks}.
\newblock \bibinfo{journal}{{\em arXiv preprint arXiv:1909.01315\/}}
  (\bibinfo{year}{2019}).
\newblock


\bibitem[\protect\citeauthoryear{Wang, Zheng, and Xue}{Wang
  et~al\mbox{.}}{2014}]%
        {wang2014travel}
\bibfield{author}{\bibinfo{person}{Yilun Wang}, \bibinfo{person}{Yu Zheng},
  {and} \bibinfo{person}{Yexiang Xue}.} \bibinfo{year}{2014}\natexlab{}.
\newblock \showarticletitle{Travel time estimation of a path using sparse
  trajectories}. In \bibinfo{booktitle}{{\em Proceedings of the 20th ACM SIGKDD
  international conference on Knowledge discovery and data mining}}.
  \bibinfo{pages}{25--34}.
\newblock


\bibitem[\protect\citeauthoryear{Yehezkel and Cohen}{Yehezkel and
  Cohen}{2012}]%
        {yehezkel2012degree}
\bibfield{author}{\bibinfo{person}{Aviv Yehezkel} {and} \bibinfo{person}{Reuven
  Cohen}.} \bibinfo{year}{2012}\natexlab{}.
\newblock \showarticletitle{Degree-based attacks and defense strategies in
  complex networks}.
\newblock \bibinfo{journal}{{\em Physical Review E\/}} \bibinfo{volume}{86},
  \bibinfo{number}{6} (\bibinfo{year}{2012}), \bibinfo{pages}{066114}.
\newblock


\bibitem[\protect\citeauthoryear{Yuan, Zheng, and Xie}{Yuan
  et~al\mbox{.}}{2012}]%
        {yuan2012discovering}
\bibfield{author}{\bibinfo{person}{Jing Yuan}, \bibinfo{person}{Yu Zheng},
  {and} \bibinfo{person}{Xing Xie}.} \bibinfo{year}{2012}\natexlab{}.
\newblock \showarticletitle{Discovering regions of different functions in a
  city using human mobility and POIs}. In \bibinfo{booktitle}{{\em Proceedings
  of the 18th ACM SIGKDD international conference on Knowledge discovery and
  data mining}}. \bibinfo{pages}{186--194}.
\newblock


\bibitem[\protect\citeauthoryear{Zhang, Jin, and Li}{Zhang
  et~al\mbox{.}}{2022}]%
        {zhang2022mirage}
\bibfield{author}{\bibinfo{person}{Jun Zhang}, \bibinfo{person}{Depeng Jin},
  {and} \bibinfo{person}{Yong Li}.} \bibinfo{year}{2022}\natexlab{}.
\newblock \showarticletitle{Mirage: an efficient and extensible city simulation
  framework (systems paper)}. In \bibinfo{booktitle}{{\em Proceedings of the
  30th International Conference on Advances in Geographic Information
  Systems}}. \bibinfo{pages}{1--4}.
\newblock


\bibitem[\protect\citeauthoryear{Zhu}{Zhu}{2005}]%
        {zhu2005semi}
\bibfield{author}{\bibinfo{person}{Xiaojin~Jerry Zhu}.}
  \bibinfo{year}{2005}\natexlab{}.
\newblock \showarticletitle{Semi-supervised learning literature survey}.
\newblock  (\bibinfo{year}{2005}).
\newblock


\end{thebibliography}
\clearpage
\appendix
\section{Appendix}\label{sec::append}

\subsection{Setting for Transferablity Experiment}\label{apend:transferablity}
We create mask graphs to represent graphs with slight perturbations. We maintain the same number of nodes but allow for changes in edges, such as deletion and addition, in the electricity network, primary road network, and coupled network. We then retrain the new node embeddings using Algorithm \ref{alg::train} and feed them into the value network of our primary agent to calculate the corresponding metrics introduced in Section \ref{metrics} during the node damaging process. Finally, we compare the results with the baseline, including CI and DE, on the newly generated topologies.
\begin{algorithm}[h]
  \caption{Retrain Progress for Mask Graph}
  \label{alg::train}
  \begin{algorithmic}[1]
    \Require
      $\mathbf{F}_{v} = \left\{\mathbf{f}_v, \forall v \in \mathcal{V}\right\}\in \mathbb{R}^{d \times |\mathcal{V}|}$;
      $\mathcal{G}^{mask}=(\mathcal{V}, \mathcal{E})$;
    \Ensure
      $\mathbf{F}_{v}^{new} \in \mathbb{R}^{d \times |\mathcal{V}|}$
    \For{epoch = 1\to $M$}
        \State embedding $ \Leftarrow$ GNN$(\mathbf{F}_{v}, \mathcal{G}^{mask})$
        \State distant $\Leftarrow \mathbf{dist}($embedding, $\mathbf{F}_{v})$
        \State reconstruction $\Leftarrow$ GNN(embedding, $\mathcal{G}^{mask})$
        \State loss $\Leftarrow$ reconstruction $+ A \times$ distant
        \State loss $\Rightarrow$ backward
    \EndFor
    
    \State \Return embedding $\Leftarrow$ GNN$(\mathbf{F}_{v}, \mathcal{G}^{mask})$
  \end{algorithmic}
\end{algorithm}
\end{document}